\documentclass[10pt,twocolumn,letterpaper]{article}

\usepackage{iccv}
\usepackage{times}
\usepackage{epsfig}
\usepackage{graphicx}
\usepackage{amsmath}
\usepackage{amssymb}

\usepackage{multirow}
\usepackage{makecell}

\usepackage[toc,page]{appendix}

\usepackage{rotating}
\usepackage{float}

\usepackage[pagebackref=true,breaklinks=true,letterpaper=true,colorlinks,bookmarks=false]{hyperref}

\iccvfinalcopy % *** Uncomment this line for the final submission

\def\iccvPaperID{3984} % *** Enter the ICCV Paper ID here

\newcommand{\supp}[1]{\textit{\textcolor{magenta}{#1}}}

\ificcvfinal\fi

\begin{document}

%%%%%%%%% TITLE
\title{SIR: Self-supervised Image Rectification via Seeing the Same Scene from Multiple Different Lenses}
\author{Jinlong Fan \and Jing Zhang \and Dacheng Tao \\
\vspace{-2mm}
\and UBTECH Sydney AI Centre, Faculty of Engineering, The University of Sydney
% \author{Jinlong Fan\\
% \and
}

\maketitle
\ificcvfinal\thispagestyle{empty}\fi

%%%%%%%%% ABSTRACT
\begin{abstract}
       Deep learning has demonstrated its power in image rectification by leveraging the representation capacity of deep neural networks via supervised training based on a large-scale synthetic dataset. However, the model may overfit the synthetic images and generalize not well on real-world fisheye images due to the limited universality of a specific distortion model and the lack of explicitly modeling the distortion and rectification process. In this paper, we propose a novel self-supervised image rectification (SIR) method based on an important insight that the rectified results of distorted images of a same scene from different lens should be the same. Specifically, we devise a new network architecture with a shared encoder and several prediction heads, each of which predicts the distortion parameter of a specific distortion model. We further leverage a differentiable warping module to generate the rectified images and re-distorted images from the distortion parameters and exploit the intra- and inter-model consistency between them during training, thereby leading to a self-supervised learning scheme without the need for ground-truth distortion parameters or normal images. Experiments on synthetic dataset and real-world fisheye images demonstrate that our method achieves comparable or even better performance than the supervised baseline method and representative state-of-the-art methods. Self-supervised learning also improves the universality of distortion models while keeping their self-consistency.
\end{abstract}

%%%%%%%%% BODY TEXT
\section{Introduction}

\begin{figure}[tbp]
\begin{center}
\includegraphics[width=0.9\linewidth]{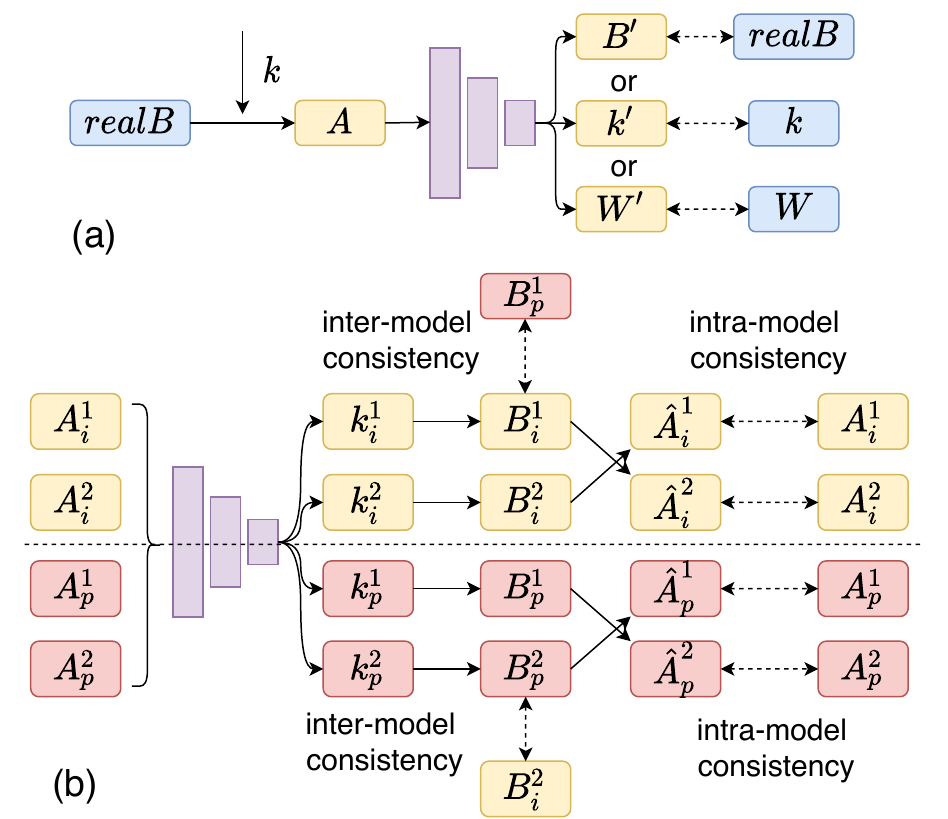}
\end{center}
  \caption{(a) Deep supervised methods predict the distortion parameters $k^\prime$, warping field $W^\prime$, or the rectified image $B^\prime$ from the distorted image $A$, which is synthesized from the normal image $realB$ based on a sampled parameter $k$ of a specific distortion model. The loss is calculated between the prediction and its ground truth, \eg, $k$ and $k^\prime$. (b) Our SIR is a deep self-supervised method without the need for any annotations. It uses the intra- and inter-model loss for training, where the former calculates the re-projection loss between the re-distorted images $\hat A$ from the same distortion model but with different parameters (marked by the same color) and the latter calculates the re-projection loss between the rectified images $B$ from different models.}
\label{fig:diff}
\end{figure}

%第一段简要介绍image rectification, 虽然进步大，但是还要很多问题

Wide field-of-view (FOV) cameras can capture wide-angle images that record more contents than conventional normal ones at a single shot, making them useful in many vision tasks \cite{markovic_MovingObjectDetection_2014,antunes_UnsupervisedVanishingPoint_2017, rituerto_VisualSLAMOmnidirectional_2010,caruso_LargescaleDirectSLAM_2015}. However, the wide FOV lenses break the pinhole camera assumption, introducing distortion in the images. To facilitate downstream applications and leverage the off-the-shelf models trained on normal images, image rectification is often used as the pre-processing step to correct the distortions. Various traditional geometry-based methods have been proposed in the past decades which formulate image rectification as an optimization problem \cite{bukhari_RobustRadialDistortion_2010,aleman-flores_AutomaticLensDistortion_2014, zhang_LinebasedMultiLabelEnergy_2015,pritts_MinimalSolversRectifying_2020}. Recently, deep learning-based methods have shown promising results by leveraging the representation power of  deep neural networks \cite{rong_RadialLensDistortion_2016, yin_FishEyeRecNetMulticontextCollaborative_2018, li_BlindGeometricDistortion_2019, liao_DRGANAutomaticRadial_2020,xue_LearningCalibrateStraight_2019, xue_FisheyeDistortionRectification_2020}. Nevertheless, how to improve the generalization of deep models that can rectify images of different distortion types and work well on real fisheye images remains challenging.

%第二段，问题一，监督方法与自监督方法，主要是对训练数据的需求，部分还需要额外的标注，如直线和场景分割才能得打比较好的效果，从合成图到实际图的适应性，参数不与视觉直接相关，很难学习，我们显示的明确了模型矫正的过程，容易学习。

Since real-world paired wide-angle and normal images are difficult to collect and no annotations of the distortion model are available, existing deep learning methods usually use synthetic datasets for training \cite{rong_RadialLensDistortion_2016,bogdan_DeepCalibDeepLearning_2018,shi_RadialLensDistortion_2018}. Wide-angle images are synthesized based on normal ones from a base dataset, \eg, ADE20k \cite{zhou_SemanticUnderstandingScenes_2019}, where a specific distortion model is used and distortion parameters are uniformly sampled from a pre-defined range. Based on the synthetic datasets, most deep learning-based methods adopt a supervised learning scheme that the network is trained to predict the distortion parameters, the equivalent warping field, or the rectified image from the distorted image as shown in Figure~\ref{fig:diff}(a). However, it is difficult to discover visual clues and learn representative features for predicting distortion parameters directly from the distorted images \cite{lopez_DeepSingleImage_2019}. To address this issue, extra annotations have been used to help the prediction. For instance,
Yin \etal~\cite{yin_FishEyeRecNetMulticontextCollaborative_2018} used scene parsing information to guide the training while Xue \etal~\cite{xue_LearningCalibrateStraight_2019} leverages annotations of straight lines to constrain the predicted distortion parameter. However, the annotations of semantic masks or geometry structures are difficult and laborious to obtain. Besides, specific network architectures or losses should be carefully devised to leverage those annotations, which may also introduce extra complexity.

%第三段，问题二，单独训练，受限于每个模型，多模型合成数据训练只是简单的累加，模型之前并无信息流动，multi-head中每个模型的能力都得到了提高，跨模型的能力得到了提升，。
The other problem is that since the training dataset is synthesized using a specific distortion model, the rectification ability of the trained network is bound with that model. It may have a good \textit{self-consistency} but with a bad \textit{universality}\footnote{The term universality is exchangeable with generalization.}, where the former refers to the ability to rectify images with the same types of distortions while the latter refers to the ability to generalize well on images from other distortion models or real-world fisheye images from different lenses \cite{tang_SelfconsistencyUniversalityCamera_2012}. Recently, some methods have been proposed to predict warping fields instead of parameters, which can represent multiple distortion models in a single framework \cite{li_BlindGeometricDistortion_2019, liao_ModelFreeDistortionRectification_2020}. The networks are trained by leveraging the supervisory signals from all the ground-truth warping fields derived from different distortion models and distortion parameters. For example, Li \etal~\cite{li_BlindGeometricDistortion_2019} used six types of distortions to generate the synthesized dataset while Liao \etal~\cite{liao_ModelFreeDistortionRectification_2020} increased the number to sixteen. However, it is difficult to regress the warping fields directly from the distorted images. Moreover, mixing all training images from different distortion models makes it hard to leverage their complementary and consistency explicitly, thereby leading to a compromised model with limited generalization. 

%对应到上面的两个问题。
In this paper, we propose a \textbf{S}elf-supervised \textbf{I}mage \textbf{R}ectification method (SIR) based on the insight that the rectified results of distorted images of the same scene from different lens should be the same. Specifically, for the distorted images from the same distortion model but with different parameters or from different distortion models (\ie, different lenses), their rectified images should be the same, \ie, \textit{intra-model consistency} in the former case and \textit{inter-model consistency} in the latter case. To exploit such consistency for image rectification, we devise a new network architecture with a shared encoder and several prediction heads, each of which predicts the distortion parameter of a specific distortion model. As shown in Figure~\ref{fig:diff}(b), the rectified images are further re-distorted by using the predicted parameters from its counterpart (marked in the same color). Then, we use a re-projection loss to calculate the difference between the input distorted image and re-distorted one (\eg, $\hat A^1_i$ and $A^1_i$) to keep intra-model consistency. Moreover, we add an inter-model consistency constraint on the rectified image from different distortion models (\eg, $B_i^1$ and $B_p^1$). In this way, SIR exploits the complementary and consistency between different distortion models and is trained in a self-supervised manner without the need for ground-truth distortion parameters or normal images. 

%贡献点
In summary, the contribution of this paper is threefold:
\begin{itemize}
    \item We propose a novel self-supervised learning method for image rectification without the need for paired training data or any annotations.
    \item We devise two novel intra- and inter-model consistency losses to leverage the complementary and consistency between different distortion models via a differentiable forward and backward warping module.
    \item Experiments on both synthetic datasets and real-world fisheye images demonstrate that our method can achieve comparable or better rectification and generalization performance than representative methods.
\end{itemize}

\section{Related Work}
%分两个小部分，一个是传统的方法，一个是深度学习的方法，提一下我们和他们的不同点。
\textbf{Image Rectification.} 
Traditional methods formulate image rectification as an optimization problem, where the objective function can be some energy and/or loss terms that measure the distortions in the image. For instance, lines are one of the most widely used visual cues since the curved lines due to distortion should be straight after rectification \cite{melo_UnsupervisedIntrinsicCalibration_2013,zhang_LinebasedMultiLabelEnergy_2015}. %Based on them, loss functions can be constructed based on geometric properties such as line equations and parallel constraints. 
Recently, some visual attention-based methods have also been proposed for image rectification \cite{carroll_OptimizingContentpreservingProjections_2009,wei_FisheyeVideoCorrection_2012,shih_DistortionfreeWideanglePortraits_2019}, which emphasize to preserve the shape of semantic content such as faces in portrait images. The attention map could be defined interactively by users or automatically by algorithms \cite{carroll_OptimizingContentpreservingProjections_2009,kim_AutomaticContentAwareProjection_2017}. Although attention map can provide informative guidance for predicting local-adaptive warping field and obtaining the better perceptual result, the optimization process is usually difficult and unstable. When multi-view images from the same lens are available, multi-view geometry constraints can also be leveraged to estimate accurate and robust distortion parameters, such as the epipolar constraint of points correspondence \cite{fitzgibbon_SimultaneousLinearEstimation_2001,sturm_MultiViewGeometryGeneral_2005,Li05anon-iterative,barreto_FundamentalMatrixCameras_2005,steele_OverconstrainedLinearEstimation_2006}. Our method shares some merits with the multi-view method but with significant differences. First, it also uses multiple images but only for training and the images are captured at the same viewpoint by different lenses (\eg, different distortion models). Second, since the distorted images in our setting share a same rectified image, dense point correspondence can be constructed via an efficient differentiable forward and backward warping module and re-projection loss is used for training, instead of sparse point correspondence and epipolar constraint-based loss in multi-view methods. 

In contrast to traditional geometry-based methods which are error-prone and time-consuming due to separate stages (\eg, line detection and rectification) and iterative optimization, efficient deep supervised learning methods have attracted increasing attention recently, including model-based methods \cite{rong_RadialLensDistortion_2016,yin_FishEyeRecNetMulticontextCollaborative_2018, bogdan_DeepCalibDeepLearning_2018, shi_RadialLensDistortion_2018, lopez_DeepSingleImage_2019,xue_LearningCalibrateStraight_2019} and model-free methods \cite{li_BlindGeometricDistortion_2019,liao_DRGANAutomaticRadial_2020,liao_ModelFreeDistortionRectification_2020}. Model-based methods directly regress the distortion parameter of a specific distortion model from the input distorted image. Their performance is limited by the universality of the distortion model. By contrast, model-free methods estimate the warping field between the distorted image and undistorted image or predict the rectified image. They can incorporate multiple models in one framework \cite{li_BlindGeometricDistortion_2019,liao_ModelFreeDistortionRectification_2020}. Our method falls into the model-based group but is a self-supervised one, which can be trained with unlabeled data. Moreover, it leverages the ``multi-view'' consistency between different distortion models with different distortion parameters, thereby improving the generalization.

\textbf{Self-supervised Learning.}
%自监督多数是将图片做几何变换，形成自监督的样本，如旋转，镜像，我们是生成了两个不同的样本
Self-supervised learning has been a hot research topic recently \cite{jing_SelfsupervisedVisualFeature_2020a,Misra_2020_CVPR,jing2020self}, which aims to learn a useful feature representation for downstream tasks by solving pretext tasks. The pretext tasks can be constructed based on spatial/temporal context, semantic labels, multi-modal correspondence, etc. For example, SimCLR defines a context-based contrasting task for self-supervised learning \cite{chen2020simple}, which obtains a comparable performance as fully supervised models. Recently, Chao \etal~\cite{chao_SelfSupervisedDeepLearning_2020} introduced self-supervised learning into image rectification, which assumes the distortion is radial-symmetric, \ie, when an image is rotated or flipped w.r.t. the center, the warping field of the distorted image should keep the same. They exploit such consistency to train the model in a self-supervised manner. By contrast, we exploit the consistency between the rectified and re-distorted images of distorted images from different distortion models, \ie, the intra-model consistency and the inter-model consistency. More generally, forward-backward consistency has been used in tracking \cite{wang_LearningCorrespondenceCycleConsistency_2019}, depth estimation \cite{yin_GeonetUnsupervisedLearning_2018, zhou_UnsupervisedLearningDepth_2017a}, image-to-image translation \cite{zhu_UnpairedImagetoImageTranslation_2017}, and text-to-image generation \cite{qiao_MirrorGANLearningTextToImage_2019} to construct self-supervision under the same transform, while the consistency exploited in our model is from different distortion models with different parameters. Beside, SIR explicitly models the distortion and rectification process via an efficient differentiable warping module, which is easy to train.

\begin{figure*}[htbp]
\begin{center}
\includegraphics[width=1\linewidth]{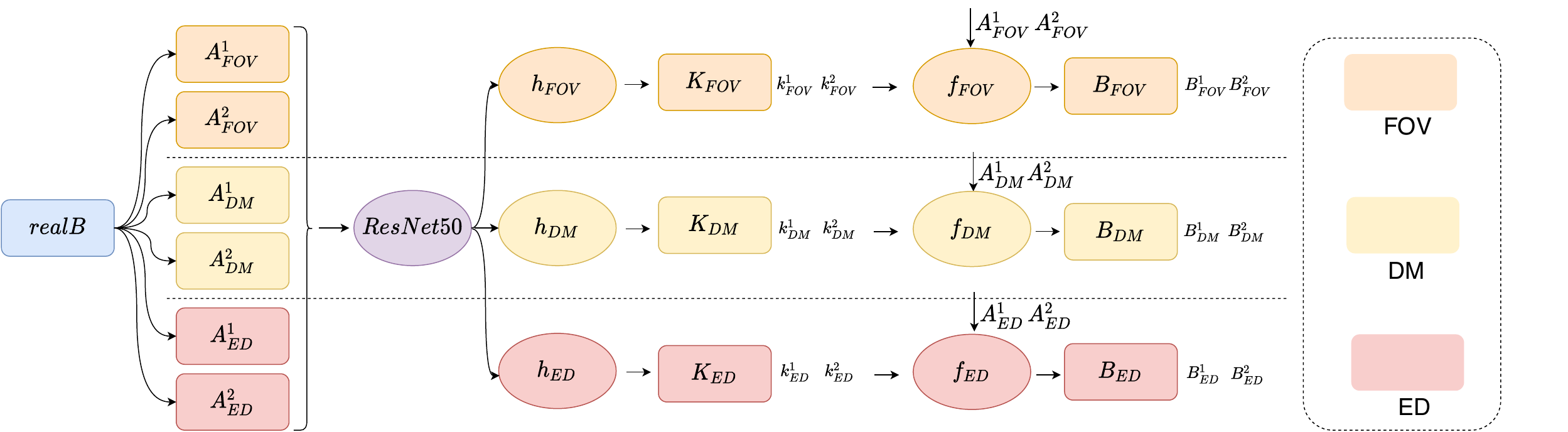}
\end{center}
  \caption{Diagram of SIR. During training, three pairs of distorted images are synthesized from a same normal image $realB$ using three distortion models, \ie, FOV, DM, and ED. The images in each pair (\eg, $A_{FOV}^1$ and $A_{FOV}^2$) are synthesized using different parameters. They are fed into a shared encoder, \ie, ResNet-50. Next, three prediction heads (\eg, $h_{FOV}$) predict the distortion parameters (\eg, $k_{FOV}^1$ and $k_{FOV}^2$) accordingly. Finally, the forward warping module (\eg, $f_{FOV}$) generates the rectified images (\eg, $B_{FOV}^1$ and $B_{FOV}^1$).}
\label{fig:archi}
\end{figure*}

\section{Self-supervised Image Rectification}
\subsection{Network Architecture}
%整体介绍网络架构，三部分，通用的，不用具体，每个部分的构成和目的，输入输出的定义  

Our self-supervised image rectification neural network has a shared encoder implemented by the ResNet-50 \cite{he_DeepResidualLearning_2016}, multiple prediction heads (\eg, three heads in this paper), and a differentiable forward and backward warping module attached to each head as illustrated in Figure~\ref{fig:archi}.

\textbf{Shared Encoder.} 
We adopt ResNet-50 as the share encoder, which embeds an input distorted image of size 257$\times$257$\times$3 into a feature vector of size 1$\times$1$\times$2048. During training, it is fed of a batch of grouped distorted images, where each group of distorted images are synthesized from a same normal image according to different distortion models with different distortion parameters. Specifically, we randomly sample two different distortion parameters from a pre-defined uniform distribution for each distortion model and synthesize two distorted images accordingly. Although there is no limit of the number of distortion models used in our method, we choose three typical ones, \ie, the \textbf{FOV} distortion model (denoted as 'FOV') \cite{tang_PrecisionAnalysisCamera_2017}, one parameter \textbf{D}ivision \textbf{M}odel (denoted as 'DM') \cite{fitzgibbon_SimultaneousLinearEstimation_2001}, and \textbf{E}qui\textbf{D}istant distortion model (denoted as 'ED') \cite{hughes_EquidistantFishEyeCalibration_2010a}. Each of them has a single parameter and an analytical forward (distorted$\rightarrow$normal) and backward (normal$\rightarrow$distorted) warping function. So totally, each group has six distorted images, \ie, $\{A_{i}^j|i\in \mathcal{M}; j=1,2\}$, $\mathcal{M} = \{FOV, DM, ED\}$, among which two for each model with different parameters.

\textbf{Prediction Head.}
We attach three prediction heads to the shared encoder, \ie, $h_{FOV}$, $h_{DM}$, and $h_{ED}$ in Figure~\ref{fig:archi}, each of which corresponds to a specific distortion model. We use a fully connected layer for each head, which has a single output neuron that outputs the predicted parameter, \ie, $K_{FOV}$, $K_{DM}$, and $K_{ED}$ in Figure~\ref{fig:archi}. Note that since the network inputs are $\{A_{i}^j|i\in \mathcal{M};j=1,2\}$, their encoded features are further split and fed into the corresponding prediction head, where each pair of parameters can be predicted as shown in Figure~\ref{fig:archi}. %They correspond to the distorted images that are generated by the same distortion model as the one modeled by the prediction head. 
Since the magnitude of the parameters from different distortion models differs significantly, we normalized the parameter to the range of $[0,1]$ and used a Sigmoid activation function after the prediction head. Specifically, assuming the minimum and maximum value of the parameter for a distortion model is $k_{min}$ and $k_{max}$, the parameter $k$ is normalized as follows:
\begin{equation}
    k = \left( k - k_{min} \right) / \left( k_{max} - k_{min} \right).
    \label{eq:k_normalize}
\end{equation}

\textbf{Warping Module.}
%distortion model的正反向传输函数
Given the predicted distortion parameter, we can get the rectified normal image using the forward warping function, which describes the mapping between corresponding pixels from the distorted image to the normal image. Conversely, given the normal image and distortion parameter, we can get the distorted image using the backward warping function. Mathematically, %we formulate these two processes as follows,
\begin{equation}
    B_i^j = f_i (A_i^j, k_i^j),
    \label{eq:forwardwarping}
\end{equation}
\begin{equation}
    \hat{A}_i^{j} = f_i^{-1} (B_i^q, k_i^j),
    \label{eq:backwardwarping}
\end{equation}
where $i\in\mathcal{M}$, $j\in\{1,2\}$, $q\in\{2,1\}$, $f_i (\cdot)$ and $f_i^{-1} (\cdot)$ represents the forward and backward warping function, respectively. $B_i^j$ is the rectified image from $A_i^j$ and $\hat{A}_i^{j}$ is the re-distorted image from $B_i^q$ using $k_i^j$. When $j\neq q$, Eq.~\eqref{eq:backwardwarping} can be used to generate a distorted image from non-paired rectified image and distortion parameter. It will be used for measure intra-model consistency, which will be detailed later. If $f_i (\cdot)$ and $f_i^{-1} (\cdot)$ have an analytical form, they can be implemented as a differentiable neural module. In this way, we can explicitly model the rectification and distortion process and implicitly set up the dense point correspondence between distorted image and normal image.

For the three models we selected, FOV and DM describe the pixel correspondence using the radial distance $r_u$ in the normal image and the radial distance $r_d$ in the distorted image, while ED sets up the pixel correspondence based on $r_d$ and the angle $\theta$ of the incident ray for each point in the distorted image. Specifically, for FOV, we have:
\begin{equation}
    r_u = f_{FOV}(r_d, k) = \frac{\tan(k r_d)}{2\tan(\frac{k}{2})},
    \label{eq:fov_fw}
\end{equation}
\begin{equation}
    r_d = f^{-1}_{FOV}(r_u, k) = \frac{1}{k} \arctan(2r_u\tan(\frac{k}{2})),
    \label{eq:fov_bw}
\end{equation}
where $r_d = \sqrt {u_d^2 + v_d^2}$, $(u_d,v_d)$ is the coordinate of a pixel on the distorted image. $r_u$ is calculated likewise. Given the pixel coordinate and distortion parameter $k$, we can obtain the coordinate of its corresponding pixel on the normal image according to Eq.~\eqref{eq:fov_fw}, and vice versa according to Eq.~\eqref{eq:fov_bw}. %Thereby, we can implement them with the ``grid\_sample'' layer in PyTorch, which is differentiable.

Compared with FOV, DM can represent relatively large distortion with fewer parameters, which has also been used widely \cite{aleman-flores_AutomaticLensDistortion_2014, rong_RadialLensDistortion_2016,shi_RadialLensDistortion_2018}. In this paper, we only use one parameter for a trade-off between complexity and accuracy. The forward warping function of DM can be written as follows,
\begin{equation}
    r_u = f_{DM}(r_d, k) = \frac{r_d}{1+k r_d^2}.
    \label{eq:DM_f}
\end{equation}
However, the backward warping function is not trivial since we can have two solutions from Eq.~\ref{eq:DM_f}, \ie,
\begin{equation}
        r_d = f^{-1}_{DM}(r_u, k) = \frac{1 \pm \sqrt{1 - 4 k r_u^2}}{2 k r_u}
\end{equation}
We choose the smaller positive value as the right solution for sampling nearby pixels. 

For the ED model, the forward and backward warping function can be written as:
\begin{equation}
    r_u = f_{ED}(\theta, k) = k \tan \theta = k \tan(\frac{r_d}{k}),
    \label{eq:ed_fw}
\end{equation}
\begin{equation}
    r_d = f^{-1}_{ED}(\theta, k) = k\theta = k\arctan(\frac{r_u}{k}).
    \label{eq:ed_bw}
\end{equation}
More details about the aforementioned distortion models and other distortion models can be found in \cite{sturm_CameraModelsFundamental_2010,tang_PrecisionAnalysisCamera_2017}.

\begin{figure}[htbp]
\begin{center}
\includegraphics[width=0.9\linewidth]{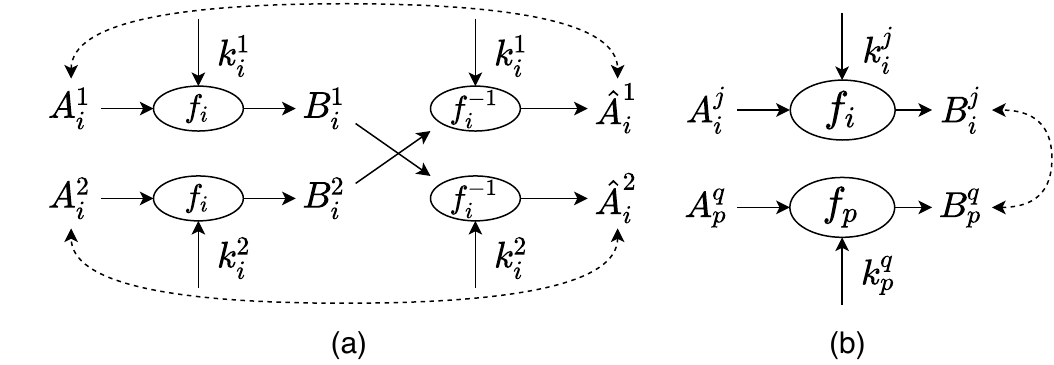}
\end{center}
  \caption{(a) The intra-model consistency loss. Given two distorted images $A_i^1$, $A_i^2$ synthesized from the same normal image using distortion model $i\in \mathcal{M}$ with different parameters $k_i^1$, $k_i^2$, the re-distorted image $\hat A_i^1$(or $\hat A_i^2$) from exchanged rectified image $B_i^2$(or $B_i^1$) should be the same with the input $A_i^1$(or $A_i^2$), \eg, $A_i^1 =\hat{A}_i^1$ and $A_i^2 =\hat{A}_i^2$. (b) The inter-model consistency loss. Given two distorted images $A_{i}^j$ and $A_{p}^q$ synthesized from the same normal image using different distortion models and parameters, their rectified images should be same, \ie, $B_{i}^j = B_{p}^q$. 
  }
\label{fig:both_loss}
\end{figure}

\subsection{Self-supervised Learning}
\label{subsec:ssl}
%loss函数定义
\textbf{Intra-model Consistency.} The intra-model consistency refers to the re-distorted images from a rectified image should be same with the corresponding input distorted images as long as they share the same distortion parameter. It is reasonable since the distorted image is uniquely determined by the parameter for one distortion model. As shown in Figure~\ref{fig:both_loss}(a), we re-distort the rectified image (\eg, $B_i^1$) with the different distortion parameter in the pair (\eg, $k_i^2$) rather than the distortion parameter (\eg, $k_i^1$) used for synthesizing the original input distorted image. In the latter case, we may obtain a trivial solution, \eg, when the prediction $k_i^1 = k_i^2 = 0$ in DM, thereby $A_i^1 = B_i^1 = \hat{A}_i^1$. By contrast, in the former case, if $k_i^1 = k_i^2 = 0$ in DM, $A_i^1 = B_i^1 = \hat{A}_i^2 \neq A_i^2$, there will be a significant difference between $\hat{A}_i^2$ and $A_i^2$, which can be used as a loss to supervise the network. Mathematically, the intra-model consistency loss can be calculated as the following L1 loss:
\begin{equation} 
    L_{intra} = \sum_{i\in \mathcal{M}}{\sum_{j=1}^{2}{| A_i^j - \hat{A}_i^j |}}.
    \label{eq:loss_intra}
\end{equation}

\textbf{Inter-model Consistency.} The inter-model consistency refers to that the rectified images of distorted images from different distortion models should be the same as long as they are synthesized from the same normal image. As shown in Figure~\ref{fig:both_loss}(b), $A_{i}^j$ and $A_{p}^q$ are two distorted images from the distortion model $i$ and $p$ with parameter $k_{i}^j$ and $k_{p}^q$, respectively. Since they are synthesized from the same normal image, thereby their rectified images should be the same, \ie, $B_{i}^j=B_{p}^q$. Mathematically, the inter-model consistency loss can be calculated as:
\begin{equation}
    L_{inter} = \sum_{(i,p)\in \hat{\mathcal{M}}}{\sum_{j=1}^{2}{\sum_{q=1}^{2}{| B_i^j - B_p^q |}}}.
    \label{eq:loss_inter}
\end{equation}
$\hat{\mathcal{M}}=\hat{\mathcal{M}}_1=\{(FOV,DM),(FOV,ED),(DM,ED)\}$, or $\hat{\mathcal{M}}=\hat{\mathcal{M}}_2=\{(FOV,DM)\}$, or $\hat{\mathcal{M}}=\hat{\mathcal{M}}_3=\{(FOV,ED)\}$, or $\hat{\mathcal{M}}=\hat{\mathcal{M}}_4=\{(DM,ED)\}$, denoting unordered combinations of all distortion models or any two models. We will present their results in the ablation study.

\textbf{Training Objective.} For self-supervised learning, the final training objective is defined as:
\begin{equation}
    L_{total} = \lambda_{intra} L_{intra} + \lambda_{inter} L_{inter}.
\end{equation}
$\lambda_{intra}$ and $\lambda_{inter}$ are hyper-parameters to balance the two losses. It is noteworthy that $L_{inter}$ cannot be used alone. Otherwise, it may lead to a trivial solution similar to what we have analyzed in the case of intra-model consistency.

%-------------------------------------------------------------------------
\section{Experiments}
\subsection{Experiment Setting}
\textbf{Dataset and Metrics.} We evaluated our method on three datasets, \ie, ADE20k~\cite{zhou_SemanticUnderstandingScenes_2019}, WireFrame~\cite{huang_LearningParseWireframes_2018}, and COCO2017~\cite{lin_MicrosoftCOCOCommon_2014}. We constructed our training and validation sets based on ADE20k training and validation sets, respectively. For WireFrame and COCO, we only used their test sets for testing since we did not train the model on these two datasets. ADE20k training set contains 20k images while the validation set contains 2k images. WireFrame test set has 462 images. We randomly taken 2k images from COCO test set as our test set. %All our training, validation and test samples were synthesized on line. 

We used three distortion models in our experiments, \ie, FOV, DM, and ED. We synthesized the dataset for each distortion model separately. Each normal image was center-cropped with the maximum size at the height or width side and then resized to $257 \times 257$. In the warping module, the image coordinate was normalized to $[-1, 1]$. The distortion parameter of each distortion model was sampled from a uniform distribution within a pre-defined range, \ie, $[-0.02, -1]$ for DM, $[0.2, 1.2]$ for FOV, and $[0.7, 2]$ for ED, which is determined according to two empirical rules. First, the distortions should not be too large such that the valid area in the distorted image would be too small. Second, distortions from different models should be comparable since we need to exploit the inter-model consistency to rectify them into the same normal image. 

For evaluation metrics, we adopt PSNR and SSIM \cite{wang_ImageQualityAssessment_2004} which have been widely used in prior arts  \cite{bukhariAutomaticRadialDistortion2013, aleman-flores_AutomaticLensDistortion_2014, rong_RadialLensDistortion_2016, yin_FishEyeRecNetMulticontextCollaborative_2018, xue_LearningCalibrateStraight_2019}. They can measure the difference between the rectified image and the ground-truth normal one.

\textbf{Implementation Detail.}
Although other backbone networks can also be used in the shared encoder, we used ResNet50 as an example in the experiment. For each mini-batch, we used 8 normal images to synthesize the distorted samples, \ie, 48 in total. We used Adam optimizer to train the network for 10 epochs on a single NVIDIA Tesla V100  GPU. The learning rate was set to 0.0001. We set $\lambda_{intra} = \lambda_{inter} = 1$ in our experiments\footnote{More details and results can be found in the \supp{supplementary material}.}. ResNet50 was initialized using the weights pre-trained on ImageNet \cite{deng_ImageNetLargescaleHierarchical_2009}. We implemented our network using PyTorch\footnote{The source code and models will be publicly available.}.

\begin{table}[ht]
\small
  \centering
    %\begin{tabular}{c c| c c c c}
     \begin{tabular}{p{0.8cm}p{0.5cm}|p{1.2cm}p{1.2cm}p{1.2cm}p{1.2cm}}
    \hline
    & & FOV & DM & ED & Avg.\\
    \hline

    \multirow{3}{*}{SL}  & FOV & \textbf{24.78/0.82}&16.57/0.45&18.67/0.54 & 20.00/0.60\\
                         & DM & 20.61/0.62&\textbf{23.51/0.78}&24.64/0.81  & \textbf{22.92/0.74}\\
                         & ED & 18.66/0.55&22.10/0.72&\textbf{24.72/0.82}  & 21.83/0.70\\
    \hline
    \multirow{3}{*}{SSL-S} & FOV & \textbf{24.26/0.80}&16.05/0.42&18.14/0.50 & 19.48/0.57\\
                          & DM & 21.52/0.67&\textbf{24.17/0.80}&\textbf{24.51/0.81}  & \textbf{23.40/0.76} \\
                          & ED & 21.13/0.67&18.86/0.58&19.26/0.60  & 19.75/0.62 \\
    \hline
    \multirow{3}{*}{SSL-M} & FOV & \textbf{23.58/0.78}&16.39/0.44&18.74/0.53 & 19.57/0.58\\
                          & DM & 21.71/0.69&\textbf{24.63/0.82}&24.77/0.81 & 23.70/0.77\\
                          & ED & 22.06/0.69&24.02/0.79&\textbf{25.48/0.84} & \textbf{23.85/0.77}\\
    \hline
    \hline
    \multirow{3}{*}{{SIR-$\hat{\mathcal{M}}_1$}} & FOV & \textbf{24.74/0.82}&16.33/0.44&18.24/0.51 & 19.77/0.59\\
                         & DM & 22.36/0.69&\textbf{24.36/0.80}&\textbf{25.29/0.83} & \textbf{24.00/0.77}\\
                         & ED & 21.47/0.66&22.64/0.73&24.23/0.79 & 22.78/0.73\\
    \hline
    \multirow{3}{*}{SIR-$\hat{\mathcal{M}}_2$} & FOV & \textbf{24.23/0.80}&16.58/0.45&18.53/0.52  & 19.78/0.59\\
                        & DM & 22.24/0.69&\textbf{24.35/0.80}&24.94/0.82  & 23.84/0.77 \\
                        & ED & 21.82/0.67&23.90/0.78&\textbf{25.98/0.85}  & \textbf{23.90/0.77}\\
    \hline             
    \multirow{3}{*}{SIR-$\hat{\mathcal{M}}_3$} & FOV & \textbf{25.54/0.84}&16.23/0.43&18.37/0.51  & 20.05/0.59\\
                        & DM & 21.54/0.66&\textbf{24.48/0.81}&24.96/0.82  & \textbf{23.66/0.76} \\
                        & ED & 20.66/0.63&23.60/0.77&\textbf{25.39/0.84}  & 23.22/0.75\\
    \hline                    
    \multirow{3}{*}{SIR-$\hat{\mathcal{M}}_4$} & FOV & \textbf{23.43/0.78}&16.23/0.44&18.45/0.52  & 19.37/0.58\\
                        & DM & 22.92/0.71&\textbf{24.90/0.83}&\textbf{25.77/0.85}  & \textbf{24.53/0.80} \\
                        & ED & 22.74/0.71&24.24/0.80&25.36/0.83  & 24.11/0.78\\
    \hline
    \end{tabular}
     \caption{PSNR and SSIM of different models on ADE20k~\cite{zhou_SemanticUnderstandingScenes_2019} validation set. For $\hat{\mathcal{M}}_i$, please refer to Section~\ref{subsec:ssl}. FOV, DM, and ED denote the \textbf{FOV} distortion model, one-parameter \textbf{D}ivision \textbf{M}odel, and the \textbf{E}qui\textbf{D}istant distortion model, respectively.}
    \label{tab:ablation_onM}
\end{table}

\subsection{Ablation Studies}
\label{subsec:exp_ablation}
\textbf{Learning Paradigm.}
%模型分成两部分，一部分是每次只要一个head单独训练，三个模型，multitask训练一个模型三个head，baseline是用GT训练，三个head单独训练，所以也有三个模型
%主要证明了方法的有效性
In order to verify the effectiveness of our self-supervised method, we set up a supervised learning baseline (denoting ``SL'') and two self-supervised variants of our SIR model (denoting ``SSL-S'' and ``SSL-M''). SL uses the ground-truth normal image to calculate the L1 loss for supervised training. It has the same architecture as our SIR model but with only a single prediction head and the forward warping module. Three networks were trained for each distortion model separately. SSL-S has the same architecture as our SIR model but with only a single prediction head and the forward and backward warping module. Intra-model consistency loss is used to train SSL-S in a self-supervised manner. Likewise, three networks were trained for each distortion model separately and used for evaluation. In contrast to SSL-S, SSL-M has the exact same architecture as our SIR model and is trained using the intra-model consistency loss. The configurations of $\hat{\mathcal{M}}_1 \sim \hat{\mathcal{M}}_4$ are defined in Section~\ref{subsec:ssl}, which denote the different choices of the inter-model loss. We evaluated these models on ADE20k validation set. The results are summarized in Table~\ref{tab:ablation_onM}. Note that the name in the first row denotes the distortion model that is used to synthesize that dataset while the name in the second column denotes the distortion model used for rectification.
%we distinguish the rectification model by the distortion model that is used to synthesize the training set for training that rectification model, as denoted by the second column in Table~\ref{tab:ablation_onM}. 
For SSL-M and SIR with multiple prediction heads, the name in the second column denotes the corresponding head.

\begin{table}[ht]
\small
  \centering
    \begin{tabular}{c| c c c c}
    \hline
    % & & \multicolumn{4}{c|}{WireFrame} & \multicolumn{4}{c}{COCO} \\
    % \cline{3-10}
     & FOV & DM & ED & Avg \\
    \hline
    FOV & \textbf{23.56/0.81}&16.23/0.50&18.66/0.59 & 19.48/0.63 \\
    DM & 23.48/0.77&\textbf{24.96/0.84}&\textbf{25.93/0.86}  & \textbf{24.97/0.82} \\
    ED & 23.20/0.76&24.46/0.82&25.57/0.85  & 24.41/0.81 \\
    \hline
    \hline
    FOV & \textbf{22.91/0.76}&15.91/0.42&18.09/0.50 & 18.97/0.56 \\
    DM & 22.48/0.70&\textbf{24.42/0.81}&\textbf{25.35/0.84} & 23.42/0.78 \\
    ED & 22.22/0.69&23.81/0.79&24.95/0.83 & \textbf{23.66/0.77} \\
    \hline
    \end{tabular}
    \caption{PSNR and SSIM of different models on WireFrame~\cite{huang_LearningParseWireframes_2018} (the first three rows) and COCO~\cite{lin_MicrosoftCOCOCommon_2014} (the last three rows) test sets. FOV, DM, and ED denote the \textbf{FOV} distortion model, one-parameter \textbf{D}ivision \textbf{M}odel, and the \textbf{E}qui\textbf{D}istant distortion model. 
  }
    \label{tab:WC_datasets}
\end{table}

From Table~\ref{tab:ablation_onM}, we have several findings. \textbf{First}, FOV and DM generally have a good self-consistency property since the model trained on images synthesized by a specific distortion model performs best on the corresponding test set synthesized by the same distortion model, as shown in the diagonal. ED also has a good self-consistency in most cases except for SSL-S, SIR-$\hat{\mathcal{M}}_1$ and SIR-$\hat{\mathcal{M}}_4$, where DM outperforms ED. As indicated by the average score across all the test sets, DM generally has the best universality in all the cases except for SSL-M and SIR-$\hat{\mathcal{M}}_2$, where DM is only marginally worse than ED. This finding is the same as that in \cite{tang_SelfconsistencyUniversalityCamera_2012}. \textbf{Second}, SSL-S achieves comparable performance as SL, which confirms the value of the proposed self-supervised learning idea. Moreover, the universality of DM by SSL-S is even better than that by SL, implying a better generalization ability for dealing with various distortions. \textbf{Third}, after employing multiple prediction heads, SSL-M improves the performance further, especially for the ED case, whose universality improves from 19.75dB to 23.85dB. Note that SSL-M did not use the inter-model consistency loss. Thereby, it demonstrates that joint training of multiple heads in a multi-task learning framework is beneficial, probably because a better shared encoder can be learned by exploiting the complementarity between different distortion models. 

\begin{figure*}[ht]
\begin{center}
\includegraphics[width=1.0\linewidth]{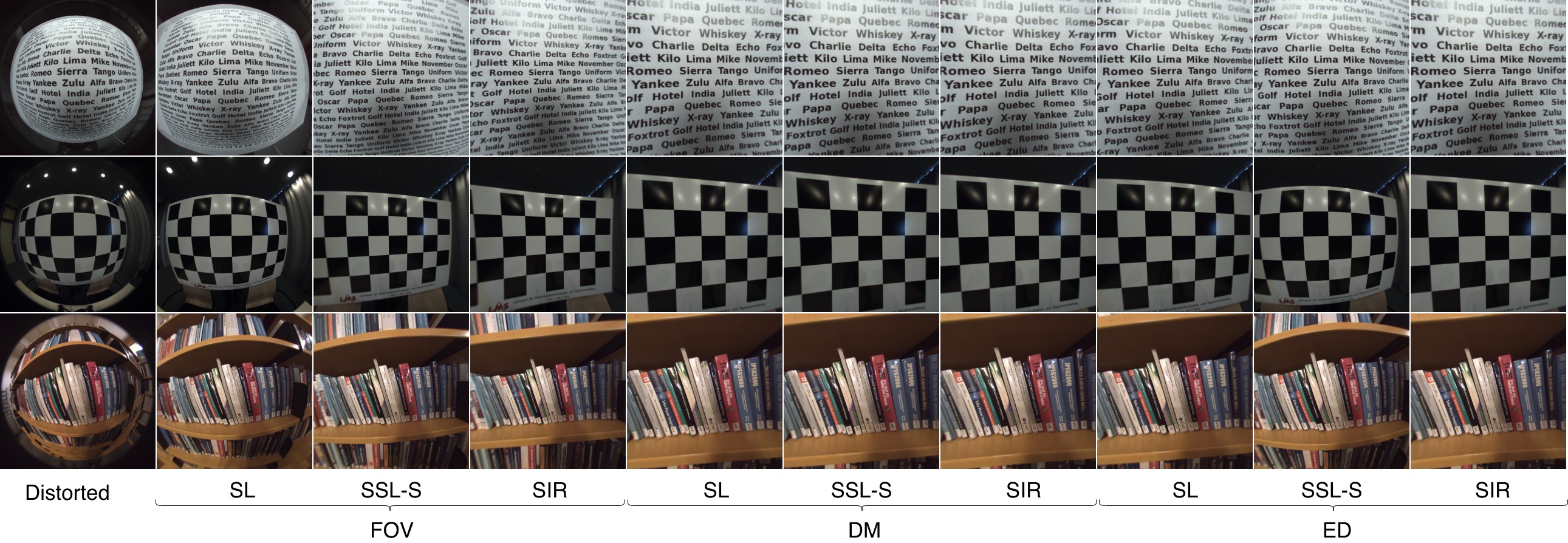}
\end{center}
  \caption{Visual comparison between the supervised baseline and several variants of our SIR model on the real fisheye video dataset \cite{eichenseer_DataSetProviding_2016}. The meaning of SL, SSL-S and SIR are detailed in the end of the first paragraph in Section~\ref{subsec:exp_ablation}.}
\label{fig:visual_self}
\end{figure*}

\textbf{Choice of the Inter-model Loss.}
From Table~\ref{tab:ablation_onM}, we also have several empirical findings about the choice of the inter-model loss. \textbf{First}, compared with SSL-M, using the inter-model consistency losses did not make significant improvement, but the universality for each model could be slightly improved in different configuration. For example, the universality of DM and ED are lifted from 23.70db to 24.53db and 23.85db to 24.11db respectively. \textbf{Second}, comparing SIR-$\hat{\mathcal{M}}_2$ and SIR-$\hat{\mathcal{M}}_3$ with SSL-M, the universality of DM and ED is not be improved significantly or even becomes worse while the universality of FOV is indeed improved, 
% \eg, from 19.57dB to 19.78dB and 20.05dB, 
showing that the prediction head of FOV may take advantage of DM and ED from the inter-model consistency. In addition, DM and ED seem to be more consistent with each other than FOV, since their universality of SIR-$\hat{\mathcal{M}}_4$ is better than that of SIR-$\hat{\mathcal{M}}_2$ and SIR-$\hat{\mathcal{M}}_3$. It can also explain why using all inter-model consistency loss in SIR-$\hat{\mathcal{M}}_1$ does not lead to better performance. In conclusion, we choose SIR-$\hat{\mathcal{M}}_4$ as the default setting due to its best universality.

Based on the default setting, we evaluated the model trained on ADE20k training set directly on WireFrame and COCO test sets without retraining or finetuning. The results are summarized in Table~\ref{tab:WC_datasets}. It can be seen that the results and trends are consistent with those on ADE20k, which demonstrate that our model has a good generalization ability. The complete results can be found in \supp{supplementary material}.

\textbf{Visual Results.}
We present some visual results obtained by the supervised learning baseline and different variants of our SIR model in Figure~\ref{fig:visual_self}. The test images are from the real fisheye video dataset~\cite{eichenseer_DataSetProviding_2016}. \textbf{First}, SL and SSL-S trained on the images synthesized based on FOV have limited ability for image rectification, although SSL-S performs a little better than SL as shown in the second and third column. It is consistent with the object metrics in Table~\ref{tab:ablation_onM} that FOV has the worst universality. Nevertheless, our SIR model performs much better than SL and SSL-S, \ie, the distortion can be corrected to a great extent as shown in the fourth column, demonstrating that the proposed self-supervised learning model based on both intra-model and inter-model consistency can improve the universality of FOV significantly. \textbf{Second}, in the case of DM, all the three models achieve comparable performance and can correct the distortion successfully. Together with the results of objective metrics, we can conclude that the DM distortion model has the best representation ability to account for different distortion types than others (\ie, best universality). \textbf{Third}, in the case of ED, the visual results are also consistent with those in Table~\ref{tab:ablation_onM} that SSL-S performs worst. In general, our SIR model has the best generalization performance on real images.

\begin{figure*}[ht]
\begin{center}
\includegraphics[width=1.0\linewidth]{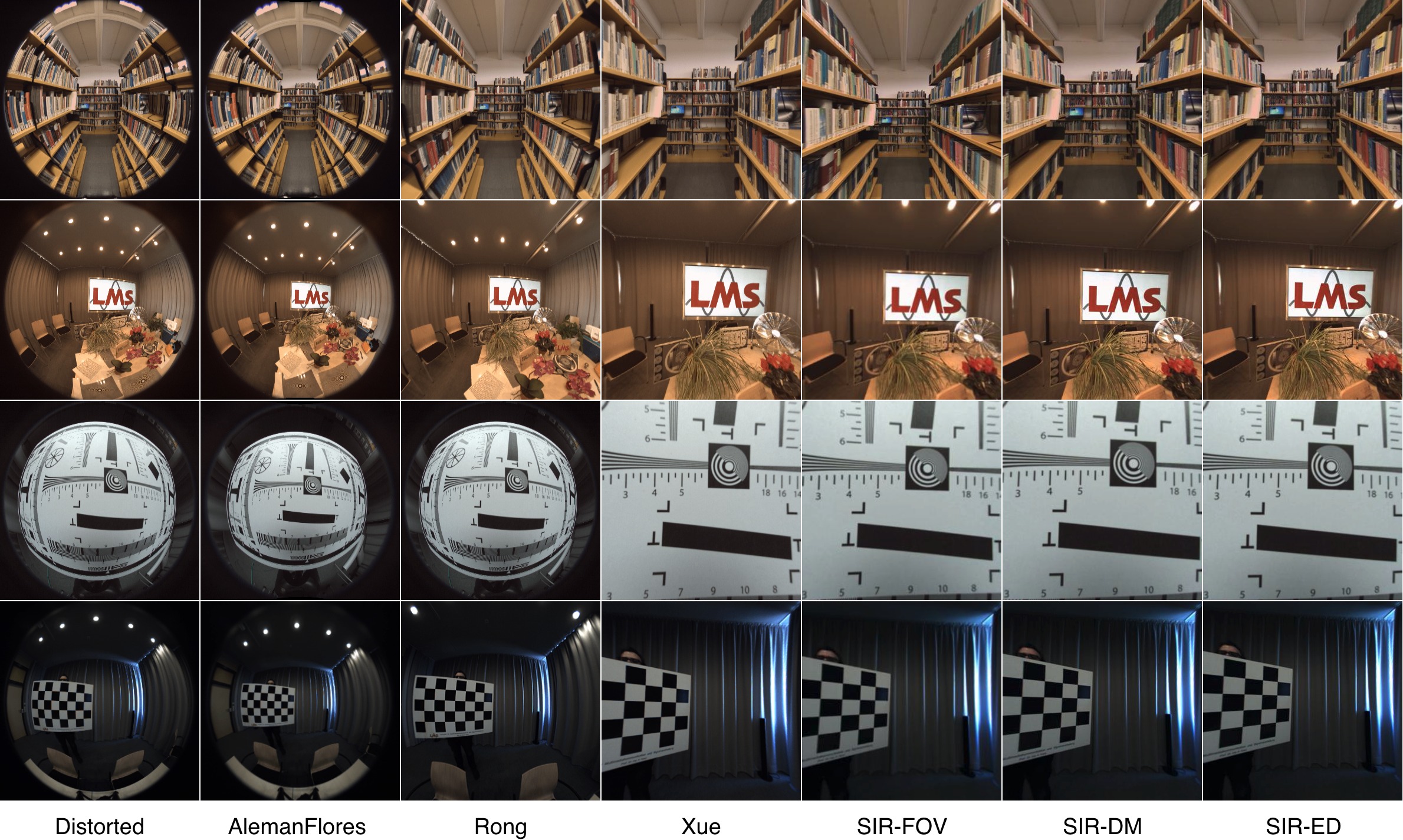}
\end{center}
  \caption{Visual comparison between SIR and several representative methods including Alem\'an-Flores~\cite{aleman-flores_AutomaticLensDistortion_2014}, Rong~\cite{rong_RadialLensDistortion_2016}, and Xue~\cite{xue_LearningCalibrateStraight_2019}, on the real fisheye video dataset \cite{eichenseer_DataSetProviding_2016}. FOV, DM, and ED denote the corresponding prediction head in our SIR model.}
\label{fig:visual_sota}
\end{figure*}

\subsection{Comparison with Representative Methods}
\label{subsec:exp_sota}

\begin{table}[ht]
\small
  \centering
    \begin{tabular}{r | c c c c c}
    \hline
    Method & Dataset & Model & Labels & PSNR & SSIM  \\
    \hline
    % Bukhar\cite{bukhariAutomaticRadialDistortion2013} & 9.34 & 0.18\\
    % Alem\'an-Flores~\cite{aleman-flores_AutomaticLensDistortion_2014} & 10.23 & 0.26\\
    % Alem\'an-Flores~\cite{aleman-flores_AutomaticLensDistortion_2014} & 13.46 & 0.49\\
    AF~\cite{aleman-flores_AutomaticLensDistortion_2014} & A &DM &- & 12.87 & 0.32  \\
    % Rong\cite{rong_RadialLensDistortion_2016} & 12.92 & 0.32\\
    % Rong~\cite{rong_RadialLensDistortion_2016} & 17.48 & 0.55\\
    Rong~\cite{rong_RadialLensDistortion_2016} & A &DM &k & 17.52 & 0.55  \\
    Yin~\cite{yin_FishEyeRecNetMulticontextCollaborative_2018} & A&Poly &\makecell[c]{image+ \\ parsing} & 14.96 & 0.41  \\
    Ours & A & DM & - & \textbf{25.77} & \textbf{0.85}  \\
    \hline
    \hline
    AF~\cite{aleman-flores_AutomaticLensDistortion_2014} & W&DM &-&13.38 &0.54  \\
    Rong~\cite{rong_RadialLensDistortion_2016} & W &DM &k & 17.53  &0.55   \\
    % \cite{xue_LearningCalibrateStraight_2019} & poly & lines+k & \textbf{27.61} & \textbf{0.87} & WF \\
    Xue~\cite{xue_LearningCalibrateStraight_2019}  & W & Poly & lines+k & \textbf{27.61} & \textbf{0.87} \\
    % \cite{chao_SelfSupervisedDeepLearning_2020} & FET & - & 15.22 & 0.498 & WML \\
    Chao~\cite{chao_SelfSupervisedDeepLearning_2020}  & W & FET & - & 12.85 & 0.35 \\
    Ours & W & Avg. & - & 24.97 & 0.82  \\
    \hline
    \hline
    AF~\cite{aleman-flores_AutomaticLensDistortion_2014}   & C &DM &- &13.08  &0.47 \\
    Rong~\cite{rong_RadialLensDistortion_2016}   & C &DM &k &17.49  &0.55 \\
    Ours  & C & DM & - & \textbf{25.35} & \textbf{0.84} \\
    \hline
    \end{tabular}
    \caption{PSNR and SSIM of SIR and representative methods. DM: one-parameter \textbf{D}ivision \textbf{M}odel. Poly: \textbf{Poly}nomial distortion model. FET: \textbf{F}ish-\textbf{E}ye \textbf{T}ransform distortion model. A: ADE20k, W: WireFrame, C: COCO.
    % F:Fish-SUNCG, M:MCindoor20000, L:LSUN Churches
    }
    \label{tab:sota_psnr_ssim}
\end{table}

We compared our SIR methods with a traditional method proposed by Alem\'an-Flores \etal in~\cite{aleman-flores_AutomaticLensDistortion_2014} and four deep learning methods proposed by Rong~\cite{rong_RadialLensDistortion_2016}, Yin~\cite{yin_FishEyeRecNetMulticontextCollaborative_2018}, Chao~\cite{chao_SelfSupervisedDeepLearning_2020} and Xue~\cite{xue_LearningCalibrateStraight_2019}. Rong's method classifies distortion parameters into 401 categories via a deep neural network and obtains the predicted parameter by weighted average during inference. Xue's method is the state-of-the-art, which first trains a line detection module using extra line annotations and then trains a rectification module to regress the distortion parameter by emphasizing the re-projection loss of lines. Scene parsing labels are used as extra information in \cite{yin_FishEyeRecNetMulticontextCollaborative_2018}. We re-implemented Rong's method by increasing the number of categories from 10 to 401 progressively and fine-tuned the model stage-by-stage. In this way, it converges faster and performs slightly better. For method that does not open source or is hard to repeat, we use the metrics published in the paper. For fairness, we try to compare the results under the same distortion model and the same dataset. If it is not possible, the average results of our model is used. PSNR and SSIM for each method are summarized in Table~\ref{tab:sota_psnr_ssim}. As can be seen, our SIR outperforms SOTA methods by a large margin across different datasets except Xue's. It is comparable with Xue's method which leverages lines to regularize the rectification process. Note that SIR does not need any annotations, thereby it has great potential in practical applications.

We also compared SIR with these methods on real fisheye images as shown in Figure~\ref{fig:visual_sota}. As can be seen, the traditional method has a limited image rectification ability on real fisheye images. Rong's method can correct the distortion to some extent, though not very pleasing. Xue's method achieves the best performance among them, which can successfully recover the normal image. Our method achieves comparable results with Xue's method no matter which prediction head is used, \eg, straight lines and upright rectangles. The results validate that our self-supervised method can efficiently leverage the intra-model and inter-model consistency and improve the universality of all the distortion models, leading to a strong generalization ability.

\subsection{Limitation Discussion and Future Work}
The proposed self-supervised method shows promising results on both synthesized distorted images and real fisheye images. Nevertheless, there is still room for further improvement. First, we only choose three typical distortion models in this paper which have analytical forward and backward warping functions. Other distortion models with the same property can be explored in our method. Besides, for models only having an analytical warping function in one direction, \ie, forward or backward, our method can be extended further to be compatible with them by predicting the warping field instead of distortion parameter. On the other hand, although we only use synthetic images for training, our method has the potential to use unlabeled real distorted images for training, which can be captured by either changing lenses or parameters such as focal length.

\section{Conclusion}
In this paper, we proposed the self-supervised learning idea for image rectification by exploiting intra-model and inter-model consistency. With a shared encoder and multiple prediction heads, our model can learn better an encoding feature representation via the complementary back-propagate signals from different heads. Both kinds of consistency improve the universality of all the three distortion models, leading to a model with better generalization ability on real fisheye images. The proposed self-supervised learning method is scalable and flexible that can be improved further by involving more distortion models with/without analytical forward and backward warping functions.

{\small
\bibliographystyle{ieee_fullname}
\bibliography{imageRec}

\begin{thebibliography}{10}\itemsep=-1pt

\bibitem{aleman-flores_AutomaticLensDistortion_2014}
Miguel {Alem{\'a}n-Flores}, Luis Alvarez, Luis Gomez, and Daniel
  {Santana-Cedr{\'e}s}.
\newblock Automatic {{Lens Distortion Correction Using One}}-{{Parameter
  Division Models}}.
\newblock {\em Image Processing On Line}, 4:327--343, 2014.

\bibitem{antunes_UnsupervisedVanishingPoint_2017}
Michel Antunes, Joao~P. Barreto, Djamila Aouada, and Bjorn Ottersten.
\newblock Unsupervised {{Vanishing Point Detection}} and {{Camera Calibration}}
  from a {{Single Manhattan Image}} with {{Radial Distortion}}.
\newblock In {\em Proceedings of {{IEEE Conference}} on {{Computer Vision}} and
  {{Pattern Recognition}}}, pages 6691--6699, 2017.

\bibitem{barreto_FundamentalMatrixCameras_2005}
J.P. Barreto and K. Daniilidis.
\newblock Fundamental matrix for cameras with radial distortion.
\newblock In {\em Proceedings of {{IEEE International Conference}} on
  {{Computer Vision Systems}}}, pages 625--632 Vol. 1, 2005.

\bibitem{bogdan_DeepCalibDeepLearning_2018}
Oleksandr Bogdan, Viktor Eckstein, Francois Rameau, and Jean-Charles Bazin.
\newblock {{DeepCalib}}: A deep learning approach for automatic intrinsic
  calibration of wide field-of-view cameras.
\newblock In {\em ACM SIGGRAPH}, 2018.

\bibitem{bukhari_RobustRadialDistortion_2010}
Faisal Bukhari and Matthew~N. Dailey.
\newblock Robust {{Radial Distortion}} from a {{Single Image}}.
\newblock In {\em Advances in {{Visual Computing}}}, volume 6454, pages 11--20,
  2010.

\bibitem{bukhariAutomaticRadialDistortion2013}
Faisal Bukhari and Matthew~N. Dailey.
\newblock Automatic {{Radial Distortion Estimation}} from a {{Single Image}}.
\newblock {\em Journal of Mathematical Imaging and Vision}, 45(1):31--45, Jan.
  2013.

\bibitem{carroll_OptimizingContentpreservingProjections_2009}
Robert Carroll, Maneesh Agrawal, and Aseem Agarwala.
\newblock Optimizing content-preserving projections for wide-angle images.
\newblock In {\em ACM SIGGRAPH}, 2009.

\bibitem{caruso_LargescaleDirectSLAM_2015}
David Caruso, Jakob Engel, and Daniel Cremers.
\newblock Large-scale direct {{SLAM}} for omnidirectional cameras.
\newblock In {\em Proceedings of {{IEEE International Conference}} on
  {{Intelligent Robots}} and {{Systems}}}, pages 141--148, 2015.

\bibitem{chao_SelfSupervisedDeepLearning_2020}
Chun-Hao Chao, Pin-Lun Hsu, Hung-Yi Lee, and Yu-Chiang~Frank Wang.
\newblock Self-{{Supervised Deep Learning}} for {{Fisheye Image
  Rectification}}.
\newblock In {\em Proceedings of {{IEEE International Conference}} on
  {{Acoustics}}, {{Speech}} and {{Signal Processing}}}, pages 2248--2252, 2020.

\bibitem{chen2020simple}
Ting Chen, Simon Kornblith, Mohammad Norouzi, and Geoffrey Hinton.
\newblock A simple framework for contrastive learning of visual
  representations.
\newblock {\em arXiv preprint arXiv:2002.05709}, 2020.

\bibitem{deng_ImageNetLargescaleHierarchical_2009}
Jia Deng, Wei Dong, Richard Socher, Li-Jia Li, Kai Li, and Li {Fei-Fei}.
\newblock {{ImageNet}}: {{A}} large-scale hierarchical image database.
\newblock In {\em Proceedings of {{IEEE Conference}} on {{Computer Vision}} and
  {{Pattern Recognition}}}, pages 248--255, 2009.

\bibitem{eichenseer_DataSetProviding_2016}
Andrea Eichenseer and Andre Kaup.
\newblock A data set providing synthetic and real-world fisheye video
  sequences.
\newblock In {\em Proceedings of {{IEEE International Conference}} on
  {{Acoustics}}, {{Speech}} and {{Signal Processing}}}, pages 1541--1545, 2016.

\bibitem{fitzgibbon_SimultaneousLinearEstimation_2001}
A.W. Fitzgibbon.
\newblock Simultaneous linear estimation of multiple view geometry and lens
  distortion.
\newblock In {\em Proceedings of {{IEEE Conference}} on {{Computer Vision}} and
  {{Pattern Recognition}}}, volume~1, pages I--125--I--132, 2001.

\bibitem{he_DeepResidualLearning_2016}
Kaiming He, Xiangyu Zhang, Shaoqing Ren, and Jian Sun.
\newblock Deep {{Residual Learning}} for {{Image Recognition}}.
\newblock In {\em Proceedings of {{IEEE Conference}} on {{Computer Vision}} and
  {{Pattern Recognition}}}, pages 770--778, 2016.

\bibitem{huang_LearningParseWireframes_2018}
Kun Huang, Yifan Wang, Zihan Zhou, Tianjiao Ding, Shenghua Gao, and Yi Ma.
\newblock Learning to {{Parse Wireframes}} in {{Images}} of {{Man}}-{{Made
  Environments}}.
\newblock In {\em Proceedings of {{IEEE Conference}} on {{Computer Vision}} and
  {{Pattern Recognition}}}, pages 626--635, 2018.

\bibitem{hughes_EquidistantFishEyeCalibration_2010a}
Ciaran Hughes, Patrick Denny, Martin Glavin, and Edward Jones.
\newblock Equidistant {{Fish}}-{{Eye Calibration}} and {{Rectification}} by
  {{Vanishing Point Extraction}}.
\newblock {\em IEEE Transactions on Pattern Analysis and Machine Intelligence},
  32(12):2289--2296, Dec. 2010.

\bibitem{jing_SelfsupervisedVisualFeature_2020a}
Longlong Jing and Yingli Tian.
\newblock Self-supervised {{Visual Feature Learning}} with {{Deep Neural
  Networks}}: {{A Survey}}.
\newblock {\em IEEE Transactions on Pattern Analysis and Machine Intelligence},
  pages 1--1, 2020.

\bibitem{jing2020self}
Longlong Jing and Yingli Tian.
\newblock Self-supervised visual feature learning with deep neural networks: A
  survey.
\newblock {\em IEEE Transactions on Pattern Analysis and Machine Intelligence},
  2020.

\bibitem{kim_AutomaticContentAwareProjection_2017}
Yeong~Won Kim, Chang-Ryeol Lee, Dae-Yong Cho, Yong~Hoon Kwon, Hyeok-Jae Choi,
  and Kuk-Jin Yoon.
\newblock Automatic content-aware projection for 360 videos.
\newblock In {\em Proceedings of IEEE International Conference on Computer
  Vision}, pages 4753--4761, 2017.

\bibitem{Li05anon-iterative}
Hongdong Li and Richard Hartley.
\newblock A non-iterative method for correcting lens distortion from nine-point
  correspondences.
\newblock In {\em Proceedings of {{IEEE International Conference}} on
  {{Computer Vision Workshops}}}, 2005.

\bibitem{li_BlindGeometricDistortion_2019}
Xiaoyu Li, Bo Zhang, Pedro~V. Sander, and Jing Liao.
\newblock Blind {{Geometric Distortion Correction}} on {{Images Through Deep
  Learning}}.
\newblock In {\em Proceedings of {{IEEE Conference}} on {{Computer Vision}} and
  {{Pattern Recognition}}}, pages 4850--4859, 2019.

\bibitem{liao_DRGANAutomaticRadial_2020}
Kang Liao, Chunyu Lin, Yao Zhao, and Moncef Gabbouj.
\newblock {{DR}}-{{GAN}}: {{Automatic Radial Distortion Rectification Using
  Conditional GAN}} in {{Real}}-{{Time}}.
\newblock {\em IEEE Transactions on Circuits and Systems for Video Technology},
  30(3):725--733, 2020.

\bibitem{liao_ModelFreeDistortionRectification_2020}
Kang Liao, Chunyu Lin, Yao Zhao, and Mai Xu.
\newblock Model-{{Free Distortion Rectification Framework Bridged}} by
  {{Distortion Distribution Map}}.
\newblock {\em IEEE Transactions on Image Processing}, 29:3707--3718, 2020.

\bibitem{lin_MicrosoftCOCOCommon_2014}
Tsung-Yi Lin, Michael Maire, Serge Belongie, James Hays, Pietro Perona, Deva
  Ramanan, Piotr Doll{\'a}r, and C.~Lawrence Zitnick.
\newblock Microsoft {{COCO}}: {{Common Objects}} in {{Context}}.
\newblock In {\em Proceedings of the {{European Conference}} on {{Computer
  Vision}}}, volume 8693, pages 740--755, 2014.

\bibitem{lopez_DeepSingleImage_2019}
Manuel Lopez, Roger Mari, Pau Gargallo, Yubin Kuang, Javier {Gonzalez-Jimenez},
  and Gloria Haro.
\newblock Deep {{Single Image Camera Calibration With Radial Distortion}}.
\newblock In {\em Proceedings of {{IEEE Conference}} on {{Computer Vision}} and
  {{Pattern Recognition}}}, pages 11809--11817, 2019.

\bibitem{markovic_MovingObjectDetection_2014}
Ivan Markovic, Francois Chaumette, and Ivan Petrovic.
\newblock Moving object detection, tracking and following using an
  omnidirectional camera on a mobile robot.
\newblock In {\em Proceedings of {{IEEE International Conference}} on
  {{Robotics}} and {{Automation}}}, pages 5630--5635, 2014.

\bibitem{melo_UnsupervisedIntrinsicCalibration_2013}
R. Melo, M. Antunes, J.P. Barreto, G. Falcao, and N. Goncalves.
\newblock Unsupervised {{Intrinsic Calibration}} from a {{Single Frame Using}}
  a '{{Plumb}}-{{Line}}' {{Approach}}.
\newblock In {\em Proceedings of {{IEEE International Conference}} on
  {{Computer Vision Systems}}}, pages 537--544, 2013.

\bibitem{Misra_2020_CVPR}
Ishan Misra and Laurens van~der Maaten.
\newblock Self-supervised learning of pretext-invariant representations.
\newblock In {\em Proceedings of IEEE Conference on Computer Vision and Pattern
  Recognition}, June 2020.

\bibitem{pritts_MinimalSolversRectifying_2020}
James Pritts, Zuzana Kukelova, Viktor Larsson, Yaroslava Lochman, and Ond{\v
  r}ej Chum.
\newblock Minimal {{Solvers}} for {{Rectifying}} from {{Radially}}-{{Distorted
  Scales}} and {{Change}} of {{Scales}}.
\newblock {\em International Journal of Computer Vision}, 128(4):950--968,
  2020.

\bibitem{qiao_MirrorGANLearningTextToImage_2019}
Tingting Qiao, Jing Zhang, Duanqing Xu, and Dacheng Tao.
\newblock {{MirrorGAN}}: {{Learning Text}}-{{To}}-{{Image Generation}} by
  {{Redescription}}.
\newblock In {\em Proceedings of {{IEEE Conference}} on {{Computer Vision}} and
  {{Pattern Recognition}}}, pages 1505--1514, June 2019.

\bibitem{rituerto_VisualSLAMOmnidirectional_2010}
Alejandro Rituerto, Luis Puig, and J.J. Guerrero.
\newblock Visual {{SLAM}} with an {{Omnidirectional Camera}}.
\newblock In {\em Proceedings of {{IEEE International Conference}} on {{Pattern
  Recognition}}}, pages 348--351, 2010.

\bibitem{rong_RadialLensDistortion_2016}
Jiangpeng Rong, Shiyao Huang, Zeyu Shang, and Xianghua Ying.
\newblock Radial {{Lens Distortion Correction Using Convolutional Neural
  Networks Trained}} with {{Synthesized Images}}.
\newblock In {\em Proceedings of the {{Asian Conference}} on {{Computer
  Vision}}}, volume 10113, pages 35--49, 2016.

\bibitem{shi_RadialLensDistortion_2018}
Yongjie Shi, Danfeng Zhang, Jingsi Wen, Xin Tong, Xianghua Ying, and Hongbin
  Zha.
\newblock Radial {{Lens Distortion Correction}} by {{Adding}} a {{Weight
  Layer}} with {{Inverted Foveal Models}} to {{Convolutional Neural Networks}}.
\newblock In {\em Proceedings of {{IEEE International Conference}} on {{Pattern
  Recognition}}}, pages 1--6, 2018.

\bibitem{shih_DistortionfreeWideanglePortraits_2019}
YiChang Shih, Wei-Sheng Lai, and Chia-Kai Liang.
\newblock Distortion-free wide-angle portraits on camera phones.
\newblock {\em ACM Transactions on Graphics}, 38(4):1--12, 2019.

\bibitem{steele_OverconstrainedLinearEstimation_2006}
R.~Matt Steele and Christopher Jaynes.
\newblock Overconstrained {{Linear Estimation}} of {{Radial Distortion}} and
  {{Multi}}-view {{Geometry}}.
\newblock In {\em Proceedings of the {{European Conference}} on {{Computer
  Vision}}}, volume 3951, pages 253--264, 2006.

\bibitem{sturm_MultiViewGeometryGeneral_2005}
P. Sturm.
\newblock Multi-{{View Geometry}} for {{General Camera Models}}.
\newblock In {\em Proceedings of {{IEEE Conference}} on {{Computer Vision}} and
  {{Pattern Recognition}}}, volume~1, pages 206--212, 2005.

\bibitem{sturm_CameraModelsFundamental_2010}
Peter Sturm.
\newblock Camera {{Models}} and {{Fundamental Concepts Used}} in {{Geometric
  Computer Vision}}.
\newblock {\em Foundations and Trends in Computer Graphics and Vision},
  6(1-2):1--183, 2010.

\bibitem{tang_SelfconsistencyUniversalityCamera_2012}
Zhongwei Tang, Rafael Grompone Von~Gioi, Pascal Monasse, and Jean-Michel Morel.
\newblock Self-consistency and universality of camera lens distortion models,
  2012.

\bibitem{tang_PrecisionAnalysisCamera_2017}
Zhongwei Tang, Rafael {Grompone von Gioi}, Pascal Monasse, and Jean-Michel
  Morel.
\newblock A {{Precision Analysis}} of {{Camera Distortion Models}}.
\newblock {\em IEEE Transactions on Image Processing}, 26(6):2694--2704, June
  2017.

\bibitem{wang_LearningCorrespondenceCycleConsistency_2019}
Xiaolong Wang, Allan Jabri, and Alexei~A. Efros.
\newblock Learning {{Correspondence From}} the {{Cycle}}-{{Consistency}} of
  {{Time}}.
\newblock In {\em Proceedings of {{IEEE Conference}} on {{Computer Vision}} and
  {{Pattern Recognition}}}, pages 2561--2571, 2019.

\bibitem{wang_ImageQualityAssessment_2004}
Z. Wang, A.C. Bovik, H.R. Sheikh, and E.P. Simoncelli.
\newblock Image {{Quality Assessment}}: {{From Error Visibility}} to
  {{Structural Similarity}}.
\newblock {\em IEEE Transactions on Image Processing}, 13(4):600--612, 2004.

\bibitem{wei_FisheyeVideoCorrection_2012}
Jin Wei, Chen-Feng Li, Shi-Min Hu, Ralph~R. Martin, and Chiew-Lan Tai.
\newblock Fisheye {{Video Correction}}.
\newblock {\em IEEE Transactions on Visualization and Computer Graphics},
  18(10):1771--1783, 2012.

\bibitem{xue_LearningCalibrateStraight_2019}
Zhucun Xue, Nan Xue, Gui-Song Xia, and Weiming Shen.
\newblock Learning to {{Calibrate Straight Lines}} for {{Fisheye Image
  Rectification}}.
\newblock In {\em Proceedings of {{IEEE Conference}} on {{Computer Vision}} and
  {{Pattern Recognition}}}, pages 1643--1651, 2019.

\bibitem{xue_FisheyeDistortionRectification_2020}
Zhu-Cun Xue, Nan Xue, and Gui-Song Xia.
\newblock Fisheye {{Distortion Rectification}} from {{Deep Straight Lines}}.
\newblock {\em arXiv:2003.11386 [cs]}, 2020.

\bibitem{yin_FishEyeRecNetMulticontextCollaborative_2018}
Xiaoqing Yin, Xinchao Wang, Jun Yu, Maojun Zhang, Pascal Fua, and Dacheng Tao.
\newblock {{FishEyeRecNet}}: {{A Multi}}-context {{Collaborative Deep Network}}
  for {{Fisheye Image Rectification}}.
\newblock In {\em Proceedings of the {{European Conference}} on {{Computer
  Vision}}}, volume 11214, pages 475--490, 2018.

\bibitem{yin_GeonetUnsupervisedLearning_2018}
Zhichao Yin and Jianping Shi.
\newblock Geonet: {{Unsupervised}} learning of dense depth, optical flow and
  camera pose.
\newblock In {\em Proceedings of {{IEEE Conference}} on {{Computer Vision}} and
  {{Pattern Recognition}}}, pages 1983--1992, 2018.

\bibitem{zhang_LinebasedMultiLabelEnergy_2015}
Mi Zhang, Jian Yao, Menghan Xia, Kai Li, Yi Zhang, and Yaping Liu.
\newblock Line-based {{Multi}}-{{Label Energy Optimization}} for fisheye image
  rectification and calibration.
\newblock In {\em Proceedings of {{IEEE Conference}} on {{Computer Vision}} and
  {{Pattern Recognition}}}, pages 4137--4145, 2015.

\bibitem{zhou_SemanticUnderstandingScenes_2019}
Bolei Zhou, Hang Zhao, Xavier Puig, Tete Xiao, Sanja Fidler, Adela Barriuso,
  and Antonio Torralba.
\newblock Semantic {{Understanding}} of {{Scenes Through}} the {{ADE20K
  Dataset}}.
\newblock {\em International Journal of Computer Vision}, 127(3):302--321,
  2019.

\bibitem{zhou_UnsupervisedLearningDepth_2017a}
Tinghui Zhou, Matthew Brown, Noah Snavely, and David~G. Lowe.
\newblock Unsupervised learning of depth and ego-motion from video.
\newblock In {\em Proceedings of {{IEEE Conference}} on {{Computer Vision}} and
  {{Pattern Recognition}}}, pages 1851--1858, 2017.

\bibitem{zhu_UnpairedImagetoImageTranslation_2017}
Jun-Yan Zhu, Taesung Park, Phillip Isola, and Alexei~A. Efros.
\newblock Unpaired {{Image}}-to-{{Image Translation Using Cycle}}-{{Consistent
  Adversarial Networks}}.
\newblock In {\em Proceedings of {{IEEE International Conference}} on
  {{Computer Vision Systems}}}, pages 2242--2251, 2017.

\end{thebibliography}
}

\newpage
\begin{appendices}
\section{Overview}
This document provides more experiments, test results, and information to the main paper. In Section~\ref{experiments}, we provide more ablation studies on the inter-model loss and the setting of $\lambda_{intra}$ and $\lambda_{inter}$. More results on WireFrame~\cite{huang_LearningParseWireframes_2018} and COCO2017~\cite{lin_MicrosoftCOCOCommon_2014} datasets are also reported. More visual results are given in Section~\ref{results} both on the synthesized dataset~\cite{zhou_SemanticUnderstandingScenes_2019} and the real fisheye video dataset~\cite{eichenseer_DataSetProviding_2016}. Finally, the runtime of our method is compared with those of  \cite{aleman-flores_AutomaticLensDistortion_2014} and \cite{rong_RadialLensDistortion_2016}.

\section{More Experiments}
\label{experiments}
\subsection{More Experiments on intra- and inter-loss}
\begin{figure}
\begin{center}
\includegraphics[width=1.0\linewidth]{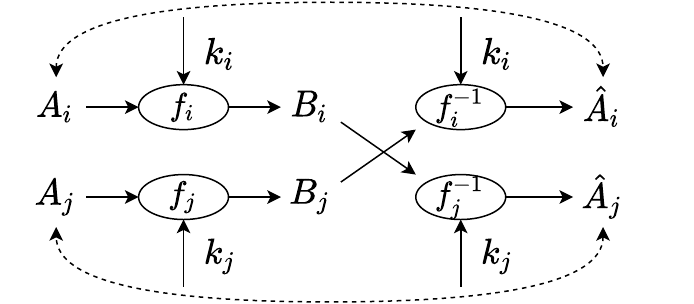}
\end{center}
  \caption{Illustration of the intra-model loss that works across different models, \ie $L_r$. $A_i$ and $A_j$ are two distorted images from the \textit{i}th and \textit{j}th distortion models. $B_i$ and $B_j$ are the rectified ones in \textit{i}th and \textit{j}th heads, respectively. $B_i$ and $B_j$ are re-distorted using the backward warping module from the other head, generating $\hat{A}_j$ and $\hat{A}_i$. }
\label{fig:intra_inter_loss}
\end{figure}

%放一些训练失败的实验，对应下文中loss部分的描述
The proposed intra-model consistent loss (denoted as $L_{intra}$) works on the distorted and re-distorted images in the same head, which means they are under the same distortion model. Here, we provide the test results when the intra-model loss is calculated between the distorted and re-distorted ones from different heads. The rectified images are re-distorted using the backward warping module in another head, as shown in Figure~\ref{fig:intra_inter_loss}. Technically, it is not an intra-model loss anymore. It is an inter-model re-projection loss. The experiment using this kind of re-projection loss is denoted as $L_r$. 

\begin{table*}[ht]
% \small
  \centering
    \begin{tabular}{c c| c c c c}
    %  \begin{tabular}{p{0.8cm}p{0.5cm}|p{1.2cm}p{1.2cm}p{1.2cm}p{1.2cm}}
    \hline
    & & FOV & DM & ED & Avg.\\
    \hline
    \multirow{3}{*}{$L_s$} & FOV & NA&NA&NA & NA\\
                         & DM & NA&NA&NA & NA\\
                         & ED & NA&NA&NA & NA\\
    \hline
    \multirow{3}{*}{$L_r$} & FOV & NA&NA&NA & NA\\
                         & DM & NA&NA&NA & NA\\
                         & ED & NA&NA&NA & NA\\
    \hline
    \multirow{3}{*}{$L_c$} & FOV & \textbf{21.80/0.71}&16.46/0.47&18.25/0.52  & 18.84/0.57\\
                        & DM & 20.68/0.66&21.50/0.68&21.27/0.67  & 21.15/0.67 \\
                        & ED & 21.09/0.67&\textbf{21.65/0.68}&\textbf{21.37/0.67}  & \textbf{21.37/0.67}\\
    \hline             
    \multirow{3}{*}{$L_s$ + $L_c$} & FOV & 11.75/0.38&11.34/0.37&11.59/0.38  & 11.56/0.38\\
                        & DM & \textbf{13.52/0.42}&\textbf{13.68/0.44}&\textbf{13.65/0.44}  & \textbf{13.62/0.43} \\
                        & ED & 13.51/0.43&13.66/0.44&13.64/0.44  & 13.60/0.44\\
    \hline                    
    \multirow{3}{*}{$L_r$ + $L_c$} & FOV & 20.82/0.68&16.42/0.46&17.69/0.50  & 18.31/0.55\\
                        & DM & 20.31/0.65&\textbf{20.14/0.62}&\textbf{19.73/0.61}  & 19.94/0.63 \\
                        & ED & \textbf{20.87/0.67}&20.02/0.61&19.69/0.60  & \textbf{20.19/0.63}\\
    \hline                    
    \multirow{3}{*}{$L_s$ + $L_r$ + $L_c$} & FOV & 20.50/0.66&16.32/0.45&17.45/0.50  & 18.09/0.54\\
                        & DM & 20.39/0.65&\textbf{19.67/0.60}&\textbf{19.22/0.59}  & \textbf{19.76/0.61} \\
                        & ED & \textbf{20.66/0.66}&19.42/0.59&19.15/0.59  & 19.74/0.61\\
    \hline
    \multirow{3}{*}{$L_{intra}$} & FOV & \textbf{23.58/0.78}&16.39/0.44&18.74/0.53 & 19.57/0.58\\
                           & DM & 21.71/0.69&\textbf{24.63/0.82}&24.77/0.81 & 23.70/0.77\\
                           & ED & 22.06/0.69&24.02/0.79&\textbf{25.48/0.84} & \textbf{23.85/0.77}\\
    \hline
    \multirow{3}{*}{$L_{intra}$ + $L_{inter}$} & FOV & \textbf{23.43/0.78}&16.23/0.44&18.45/0.52  & 19.37/0.58\\
                        & DM & 22.92/0.71&\textbf{24.90/0.83}&\textbf{25.77/0.85}  & \textbf{24.53/0.80} \\
                        & ED & 22.74/0.71&24.24/0.80&25.36/0.83  & 24.11/0.78\\
    \hline
    \end{tabular}
     \caption{More ablation studies on the inter-model loss using PSNR and SSIM. NA means the experiment fails and images can not be rectified. The setting of each expriment can be founnd in Section~\ref{experiments}.}
    \label{tab:ablation_interloss}
\end{table*}

\begin{table*}[t]
  \centering
    \begin{tabular}{c c|| c c c c | c c c c}
    \hline
    & & \multicolumn{4}{c|}{WireFrame} & \multicolumn{4}{c}{COCO2017} \\
    \cline{3-10}
    & & FOV & DM & ED & Avg & FOV & DM & ED & Avg \\
    \hline
    \multirow{3}{*}{SL} & FOV & \textbf{24.98/0.83}&16.44/0.51&18.77/0.61 & 20.06/0.65  & \textbf{24.30/0.80}&16.12/0.44&18.48/0.53 & 19.63/0.59\\
                        & DM & 20.69/0.67&\textbf{23.44/0.79}&\textbf{24.53/0.82} &\textbf{22.89/0.76} & 20.24/0.61&\textbf{23.04/0.76}&\textbf{24.25/0.80} & \textbf{22.51/0.72}\\
                        & ED & 18.77/0.62&22.30/0.75&24.47/0.82  & 21.85/0.73   & 18.41/0.55&21.59/0.71&24.18/0.80 & 21.39/0.69\\
    \hline
    \multirow{3}{*}{SSL-S} & FOV & \textbf{24.45/0.83}&16.07/0.49&18.49/0.58 & 19.67/0.63 & \textbf{23.69/0.79}&15.70/0.41&17.79/0.49 & 19.06/0.56\\
                          & DM & 21.70/0.72&\textbf{24.21/0.82}&\textbf{24.44/0.82}  & \textbf{23.45/0.79} & 21.11/0.66&\textbf{23.72/0.79}&\textbf{24.17/0.80} & \textbf{23.00/0.75} \\
                          & ED & 21.58/0.73&19.00/0.64&19.57/0.66  & 20.05/0.68  & 20.44/0.65&18.36/0.56&18.83/0.59 & 19.21/0.60 \\
    \hline
    \multirow{3}{*}{SSL-M} & FOV & \textbf{23.68/0.81}&16.07/0.49&18.83/0.60 & 19.53/0.63  & \textbf{23.58/0.78}&15.97/0.43&18.55/0.53 & 19.37/0.58 \\
                          & DM & 21.95/0.74&\textbf{24.48/0.83}&24.59/0.83  & 23.67/0.80 & 21.50/0.69&\textbf{24.34/0.81}&24.74/0.82 & 23.53/0.77 \\
                          & ED & 22.95/0.75&24.11/0.81&\textbf{26.00/0.87}  & \textbf{24.35/0.81} & 22.12/0.69&23.95/0.79&\textbf{25.71/0.85} & \textbf{23.93/0.78} \\
    \hline
    \hline
    \multirow{3}{*}{SIR-$\hat{\mathcal{M}}_1$} &FOV &\textbf{24.93/0.84} &16.17/0.50 &18.32/0.58 &19.80/0.64   &\textbf{24.44/0.81} &15.78/0.42 &17.98/0.50 &19.40/0.58 \\
    &DM &22.57/0.73 &\textbf{24.34/0.82} &\textbf{25.36/0.85} &\textbf{24.09/0.80}   &21.86/0.67 &\textbf{23.84/0.79} &\textbf{24.89/0.82} &\textbf{23.53/0.76} \\
    &ED &21.57/0.71 &22.69/0.76 &24.42/0.82 &22.89/0.76   &21.01/0.65 &22.22/0.72 &23.87/0.78 &22.37/0.72 \\
    \hline
    \multirow{3}{*}{SIR-$\hat{\mathcal{M}}_2$}  &FOV &\textbf{24.43/0.83} &16.52/0.51 &18.70/0.59 &19.88/0.64  &\textbf{24.38/0.81} &16.23/0.44 &18.28/0.52 &19.63/0.59    \\
    &DM &22.02/0.73 &\textbf{24.51/0.53} &25.10/0.84 &23.88/0.70   &21.53/0.66 &\textbf{24.12/0.80} &25.08/0.83 &\textbf{23.58/0.76} \\
    &ED &21.77/0.71 &24.09/0.81 &\textbf{25.99/0.87} &\textbf{23.95/0.80}   &20.21/0.61 &23.35/0.77 &\textbf{25.16/0.84} &22.91/0.74 \\
    \hline
    \multirow{3}{*}{SIR-$\hat{\mathcal{M}}_3$} &FOV &\textbf{25.62/0.86} &16.27/0.50 &18.40/0.58 &20.10/0.65   &\textbf{25.12/0.83} &15.83/0.42 &18.12/0.51 &19.69/0.59 \\
    &DM &21.54/0.71 &\textbf{24.52/0.83} &25.04/0.84 &\textbf{23.70/0.79}   &21.16/0.65 &\textbf{24.10/0.81} &24.88/0.83 &\textbf{23.38/0.76} \\
    &ED &20.79/0.68 &23.81/0.80 &\textbf{25.50/0.85} &23.37/0.78   &20.11/0.61 &23.21/0.76 &\textbf{25.09/0.83} &22.80/0.73 \\
    \hline
    \multirow{3}{*}{SIR-$\hat{\mathcal{M}}_4$}   & FOV & \textbf{23.56/0.81}&16.23/0.50&18.66/0.59 & 19.48/0.63  &         \textbf{22.91/0.76}&15.91/0.42&18.09/0.50 & 18.97/0.56 \\
                          & DM & 23.48/0.77&\textbf{24.96/0.84}&\textbf{25.93/0.86}  & \textbf{24.97/0.82} & 22.48/0.70&\textbf{24.42/0.81}&\textbf{25.35/0.84} & 23.42/0.78 \\
                          & ED & 23.20/0.76&24.46/0.82&25.57/0.85  & 24.41/0.81 & 22.22/0.69&23.81/0.79&24.95/0.83 & \textbf{23.66/0.77} \\
    \hline
    \end{tabular}
    \caption{PSNR and SSIM of different models tested on WireFrame~\cite{huang_LearningParseWireframes_2018} and COCO2017~\cite{lin_MicrosoftCOCOCommon_2014} datasets. 
  %FOV, DM and ED denote the \textbf{FOV} distortion model, one-parameter \textbf{D}ivision \textbf{M}odel, and the \textbf{E}qui\textbf{D}istant distortion model, respectively. GT means deep models are trained using the \textbf{G}round \textbf{T}ruth. CP means \textbf{C}ross \textbf{P}arameter models. MT means the \textbf{M}ulti-\textbf{T}ask model.
  }
    \label{tab:othertwo_datasets}
\end{table*}

\begin{figure}
\begin{center}
\includegraphics[width=1.0\linewidth]{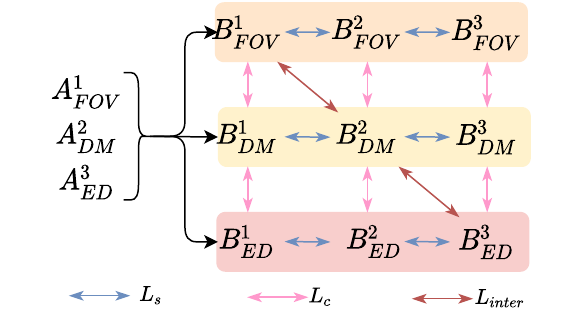}
\end{center}
  \caption{The difference between $L_s$, $L_c$ and $L_{inter}$.}
\label{fig:comp_loss}
\end{figure}

Similarly, the proposed inter-model consistent loss (denoted as $L_{inter}$) only works on the images that are rectified by the head under the same distortion model as the one used to synthesize the input images. Here, we provide the test results when the inter-model loss is added on all images both from the same head (denoted as $L_s$) and from different heads (denoted as $L_c$). The former one means that the rectified results of distorted images under different models in one specific head (\ie using one specific distortion model) should be consistent. And the latter one says the rectified outputs of different heads should be the same. To be more specific, we take FOV, DM, and ED models in the main paper as examples. The difference between $L_s$, $L_c$ and the proposed $L_{inter}$ is illustrated in Figure~\ref{fig:comp_loss}. Given three images $A_{FOV}^1$, $A_{DM}^2$, $A_{ED}^3$ from FOV, DM and ED distortion model, the rectified images in each head are $B^1_i$, $B^2_i$, $B^3_i$ ($i \in \{FOV, DM, ED\}$) respectively, \eg $B^3_{FOV}$ is the rectified result of $A^3_{ED}$ in FOV head. $L_s$ adds regulation on the results in each head, \eg $B^1_{FOV}$, $B^2_{FOV}$, $B^3_{FOV}$ in FOV head, while $L_c$ adds regulation between the results of every two heads, \eg ($B^1_{FOV}$, $B^2_{FOV}$, $B^3_{FOV}$) and ($B^1_{DM}$, $B^2_{DM}$, $B^3_{DM}$) from FOV and DM heads. Our proposed inter-model loss only works on the diagonal results, which are the rectified results of input images having the same distortion model as that in the head, \eg $B^1_{FOV}$ in FOV head for $A^1_{FOV}$, $B^2_{DM}$ in DM head for $A^2_{DM}$, and $B^3_{ED}$ in ED head for $A^3_{ED}$.

The test results are listed in Table~\ref{tab:ablation_interloss}. When $L_s$ and $L_r$ work alone, they fail to rectify the input distorted images since the images collapse to some common distorted states that all images could be transformed to. The rectified results are still distorted when the loss is minimized. If $L_c$ is used, the common state could be closer to the correct one and higher PSNR and SSIM are achieved. When $L_s$ and $L_r$ work with $L_c$, PSNR and SSIM do not be improved. On the contrary, negative effects are observed. Especially, when $L_s$ works with $L_c$ which is equivalent to regular all the rectified outputs of all input images from all heads, it gives the worst objective metrics. With the proposed $L_{intra}$ only, we can get superior performance. And together with $L_{inter}$ the highest overall PSNR and SSIM are achieved.

\subsection{Ablation Study on Loss Weights}
\begin{figure}
\begin{center}
\includegraphics[width=1.0\linewidth]{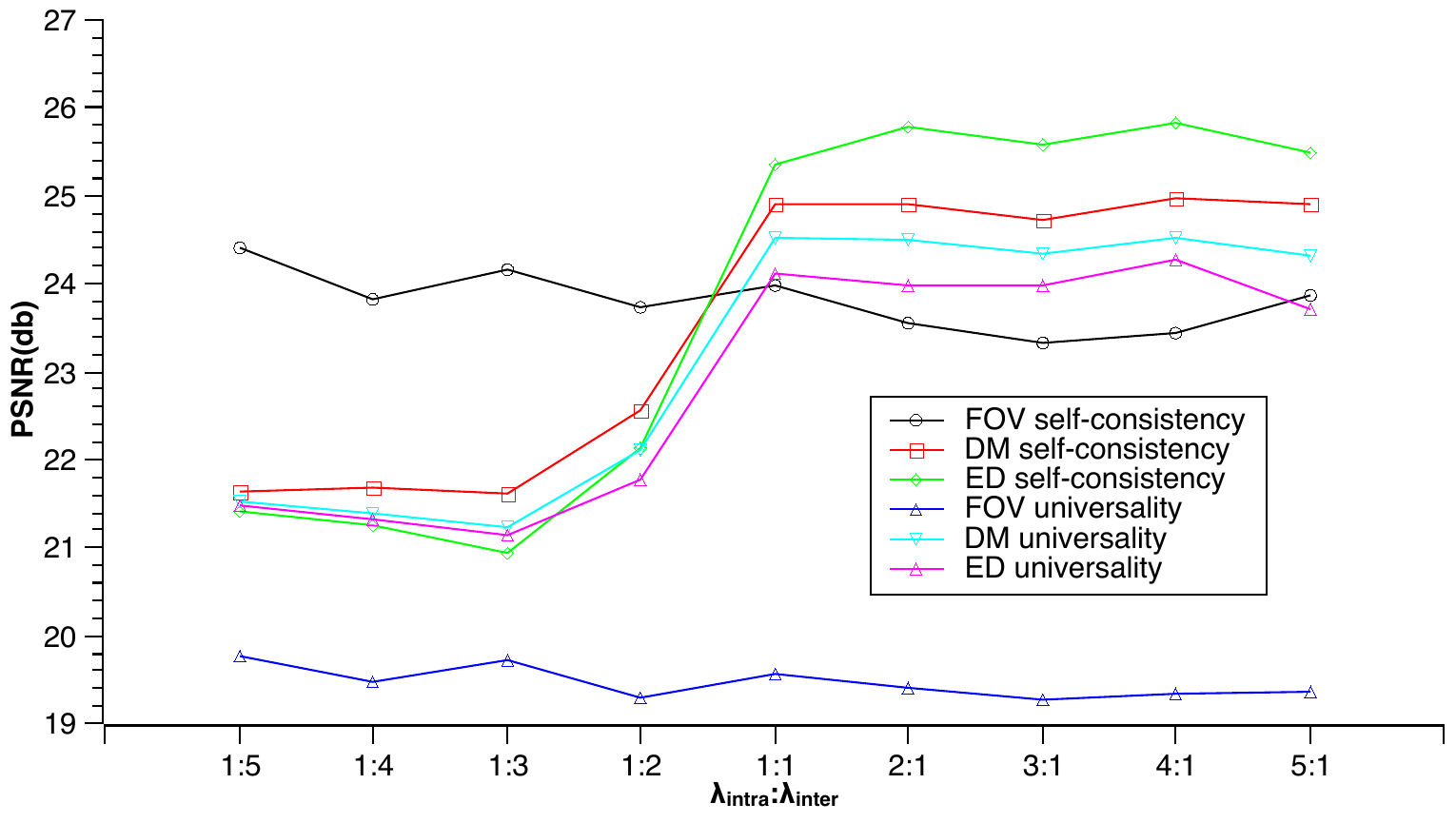}
\end{center}
  \caption{Comparison of self-consistency and universality of each model under different settings of $\lambda_{intra} : \lambda_{inter}$.}
\label{fig:loss_weights}
\end{figure}

When both $L_{intra}$ and $L_{inter}$ are exploited in training, we use two hyper-parameters $\lambda_{intra}$ and $\lambda_{inter}$ to balance them. In ablation study, we explore nine settings of the ratio between $\lambda_{intra}$ and $\lambda_{inter}$, as illustrated in Figure \ref{fig:loss_weights}. We find that $\lambda_{intra} : \lambda_{inter} = {1:1}$ is a critical point and when $\lambda_{intra} \geq \lambda_{inter}$, the training is robust to the ratio. We set  $\lambda_{intra} = \lambda_{inter} = 1$ in all our experiments.

\subsection{More Results on WireFrame and COCO}
After the model is trained on ADE20k~\cite{zhou_SemanticUnderstandingScenes_2019} dataset, we directly test it on WireFrame~\cite{huang_LearningParseWireframes_2018} and COCO2017~\cite{lin_MicrosoftCOCOCommon_2014} datasets without re-training or finetuning. The results are summarized in Table \ref{tab:othertwo_datasets}. Generally, the test results and trends are consistent with those on ADE20k, which implicitly validate the generalization and robustness of our method.

\section{Additional Visual Results}
\label{results}
%合成数据上的结果，再挑一些Fisheye dataset上的结果图
%multiscale 测试的完整结果图是否需要？

\begin{sidewaysfigure*}
\centering
\includegraphics[width=0.95\textwidth]{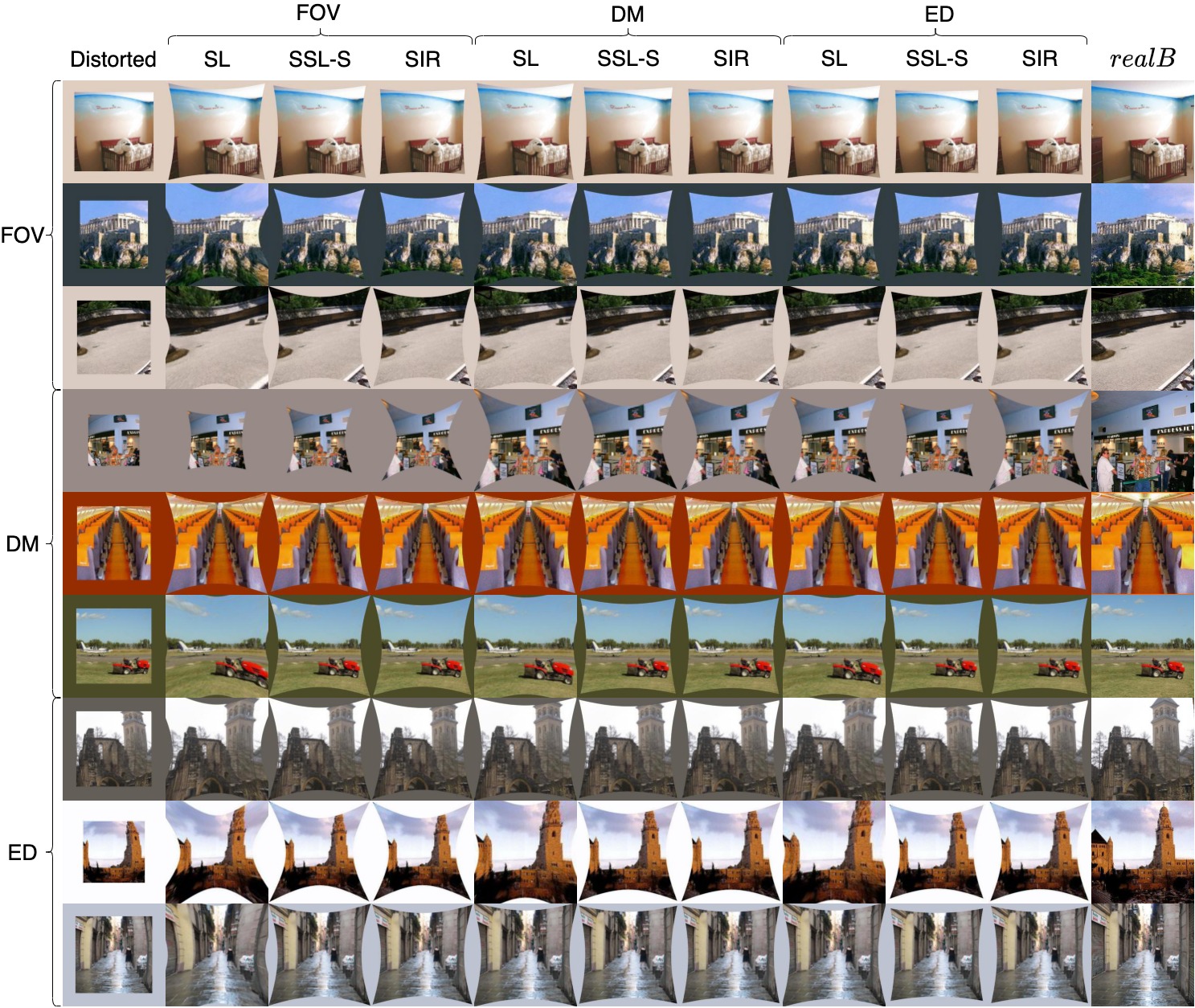}
\caption{Comparison of the visual results with the baseline methods on synthesized test dataset. The first column is the synthesized distorted image. $realB$ is the original normal image from the ADE20K test dataset~\cite{zhou_SemanticUnderstandingScenes_2019}. %The middle nine columns represents the rectified results of different models in different methods. The distorted images on 1-3 rows are the rectified results of FOV model, while 4-6 and 7-9 are those of DM model and ED model respectively.
}
\label{fig:comp_syn}
\end{sidewaysfigure*}

\begin{sidewaysfigure*}
\centering
\includegraphics[width=1.0\linewidth]{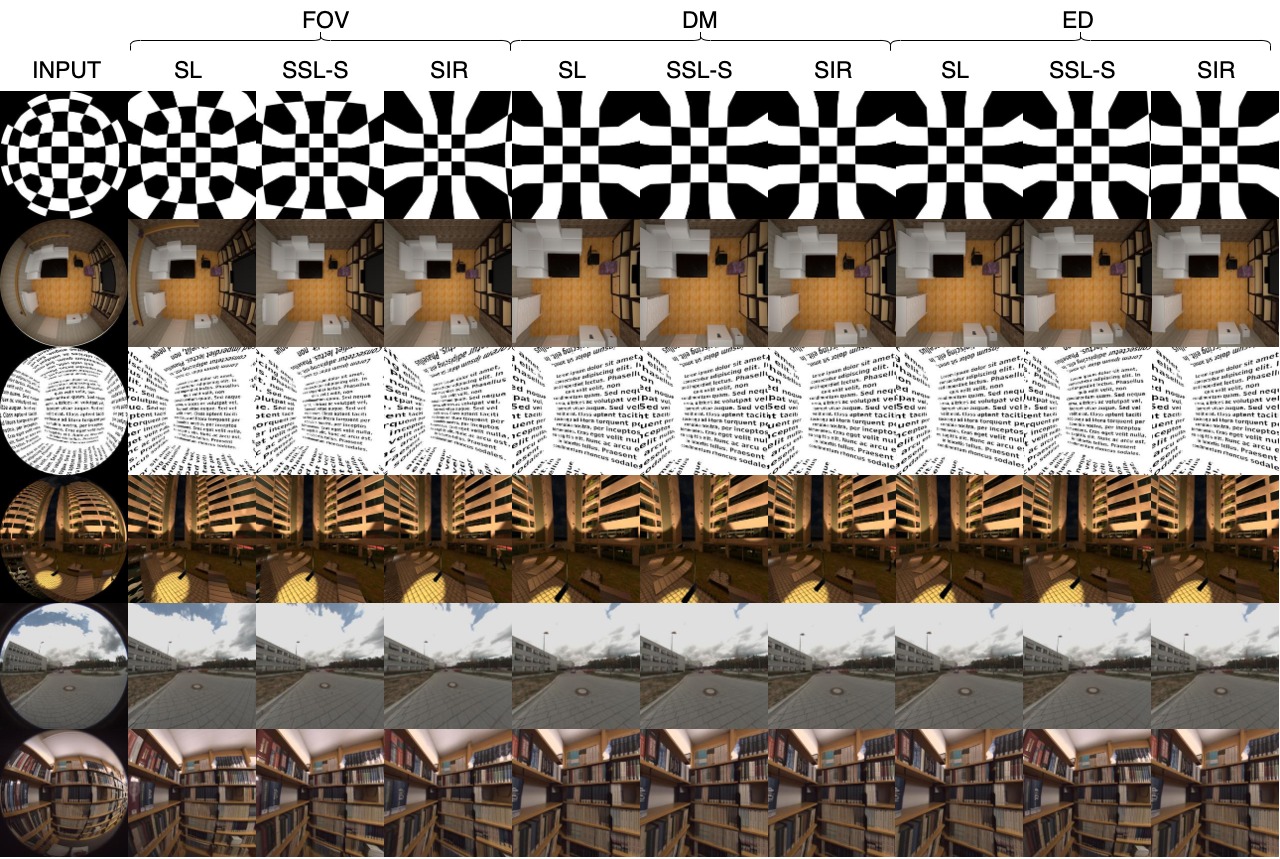}
\caption{Comparison of the visual results with the baseline methods on fisheye video dataset~\cite{eichenseer_DataSetProviding_2016}. The first column contains the input real fisheye images.}
\label{fig:comp_real}
\end{sidewaysfigure*}

More test results on the synthesized test dataset and real fisyeye video dataset~\cite{eichenseer_DataSetProviding_2016} are provided in Figure~\ref{fig:comp_syn} and Figure~\ref{fig:comp_real}. SL, SSL-S and SIR have the same meaning as in main paper.

% \begin{figure*}
% \end{figure*}

% \section{Effects of Input Pattern}
%输入pattern不同效果不同

% \section{Finetuning on Real Images}
% \label{finetune}
%

\section{Running Time}
% \label{runtime} 1.68, 1.360/1.367, 1.403/1.408/1.411/1.413
% 0.001007149089127779 0.0034005089290440084
% 0.006037094897061736 0.009365183231069813 0.01229513923277208 0.014652178571302087

We compare the running time of our method with those of the representative traditional method~\cite{aleman-flores_AutomaticLensDistortion_2014} and the parameter regression method proposed in ~\cite{rong_RadialLensDistortion_2016}. We test the images on NVIDIA Tesla V100 GPU with input size $257\times257$. Generally, traditional methods are slower than deep learning-based methods since the latter normally only need a forward pass to predict the parameters, while the former has to estimate the parameters via a time-consuming optimization procedure. Although our method is slower than Rong \etal ~\cite{rong_RadialLensDistortion_2016}, it can still run in real-time ($>60$ FPS) even with three rectification heads. 

\begin{table}[H]
% \scriptsize
    \centering
    \begin{tabular}{ c r l}
    \hline
    \multicolumn{2}{r}{Methods} & Time \\
    \hline
    % \cite{bukhariAutomaticRadialDistortion2013} & Intel i5-4200U CPU & 62.53  \\
    \multicolumn{2}{r}{Alem\'an-Flores~\cite{aleman-flores_AutomaticLensDistortion_2014}} & 1.301   \\
    % \citep{zhang_LinebasedMultiLabelEnergy_2015} & Intel i5-4200U CPU & 80.07  \\
    \hline
    \multicolumn{2}{r}{Rong~\cite{rong_RadialLensDistortion_2016}} & \textbf{0.003}  \\
    % \cite{yin_FishEyeRecNetMulticontextCollaborative_2018} & NVIDIA Tesla K80 GPU & 1.31  \\
    % \cite{liao_DRGANAutomaticRadial_2020}  & NVIDIA TITAN X GPU & 0.038  \\
    \hline 
    \multirow{3}{*}{Ours} & w/ one head  & 0.009  \\
                          & w/ two heads & 0.012  \\
                          & w/ three heads & 0.015  \\
    \hline
    \end{tabular}
    \caption{Comparison of running time (seconds).}
    \label{tab:runtime}
\end{table}

\end{appendices}

\end{document}